\documentclass{article}
\usepackage{microtype}
\usepackage{graphicx}
\usepackage{subfigure}
\usepackage{booktabs}
\usepackage{xcolor}
\usepackage{ulem}
\usepackage{threeparttable}
\usepackage{multirow}
\usepackage{caption}
\usepackage{natbib}
\usepackage{hyperref}

\usepackage[accepted]{icml2025}
\usepackage{amsmath}
\usepackage{amssymb}
\usepackage{mathtools}
\usepackage{amsthm}
\usepackage[capitalize,noabbrev]{cleveref}
\theoremstyle{plain}

\theoremstyle{definition}

\theoremstyle{remark}

\usepackage[textsize=tiny]{todonotes}
\icmltitlerunning{Gateformer}

\begin{document}
\twocolumn[
\icmltitle{Gateformer: Advancing Multivariate Time Series Forecasting via Temporal and Variate-Wise Attention with Gated Representations}
\begin{icmlauthorlist}
\icmlauthor{Yu-Hsiang Lan}{nyu}
\icmlauthor{Eric K. Oermann}{nyu,langone}
\end{icmlauthorlist}
\icmlaffiliation{nyu}{New York University, New York, NY, USA}
\icmlaffiliation{langone}{Department of Neurosurgery, NYU Langone Health, New York, NY, USA}
\icmlcorrespondingauthor{Yu-Hsiang Lan}{yl12081@nyu.edu}
\icmlkeywords{Machine Learning, ICML}
\vskip 0.3in
]
\printAffiliationsAndNotice{~}
\begin{abstract}
Transformer-based models have recently shown promise in time series forecasting, yet effectively modeling multivariate time series remains challenging due to the need to capture both temporal (cross-time) and variate (cross-variate) dependencies. While prior methods attempt to address both, it remains unclear how to optimally integrate these dependencies within the Transformer architecture for both accuracy and efficiency. We re-purpose the Transformer to explicitly model these two types of dependencies: first embedding each variate independently to capture temporal dynamics, then applying attention over these embeddings to model cross-variate relationships. Gating mechanisms in both stages regulate information flow, enabling the model to focus on relevant features. Our approach achieves state-of-the-art performance on 13 real-world datasets and can be integrated into Transformer-, LLM-based, and foundation time series models, improving performance by up to 20.7\%. Code is available at this repository: \href{https://github.com/nyuolab/Gateformer}{https://github.com/nyuolab/Gateformer}.
\end{abstract}

\section{Introduction}
Time series forecasting plays a critical role in domains such as traffic \citep{lv2014traffic}, energy \citep{zhu2023energy}, weather \citep{angryk2020multivariate}, healthcare \citep{kaushik2020ai}, and finance \citep{chen2012bayesian}. In multivariate settings, accurate forecasting depends on modeling both intra-series (temporal) and inter-series (cross-variable) dependencies \citep{Crossformer}. Transformers \citep{Transformer}, originally successful in NLP, have been adapted for time series due to their ability to model long-range dependencies \citep{wen2022transformers}.

Many Transformer-based models \citep{Reformer, Informer, Pyraformer} embed all variables at each time step into a single token and apply attention across time. However, these tokens often lack semantic depth, weakening temporal modeling—sometimes allowing simple linear models to outperform them \citep{DLinear}. PatchTST \citep{PatchTST} addresses this by using patch-level temporal attention but overlooks cross-variable dependencies. iTransformer \citep{iTransformer} models cross-variable correlations via coarse embeddings, sacrificing fine-grained temporal detail.

We propose \textbf{Gateformer}, a Transformer-based model that explicitly captures both temporal and cross-variate dependencies. Each variate is first encoded independently to capture temporal patterns, then integrated with global temporal context through a gating mechanism that regulates information flow. These representations serve as inputs for cross-variate modeling.

To model cross-variate dependencies, we apply self-attention over variate representations, producing variate-interacting embeddings. While such modeling improves capacity \citep{han2024capacity}, it may degrade performance on low-dimensional datasets. To address this, we fuse variate-interacting and non-interacting embeddings through a second gating mechanism, dynamically controlling cross-variate influence and ensuring robust performance across dataset scales. Our main contributions in this work can be summarized as follows:
\begin{itemize}
    \item We introduce Gateformer, which combines temporal and variate-wise attention with gated representations to improve multivariate time series forecasting.
    \item Gateformer achieves state-of-the-art results on 13 real-world benchmarks, ranking top-1 in 91 and top-2 in 122 out of 130 settings.
    \item Our propsed framework seamlessly integrates with Transformer-based, LLM-based, and foundation time series models, achieving performance improvements of up to 20.7\% and facilitating the development of foundational multivariate models.
\end{itemize}

\begin{figure*}[t]
  \centering
\includegraphics[width=\textwidth]{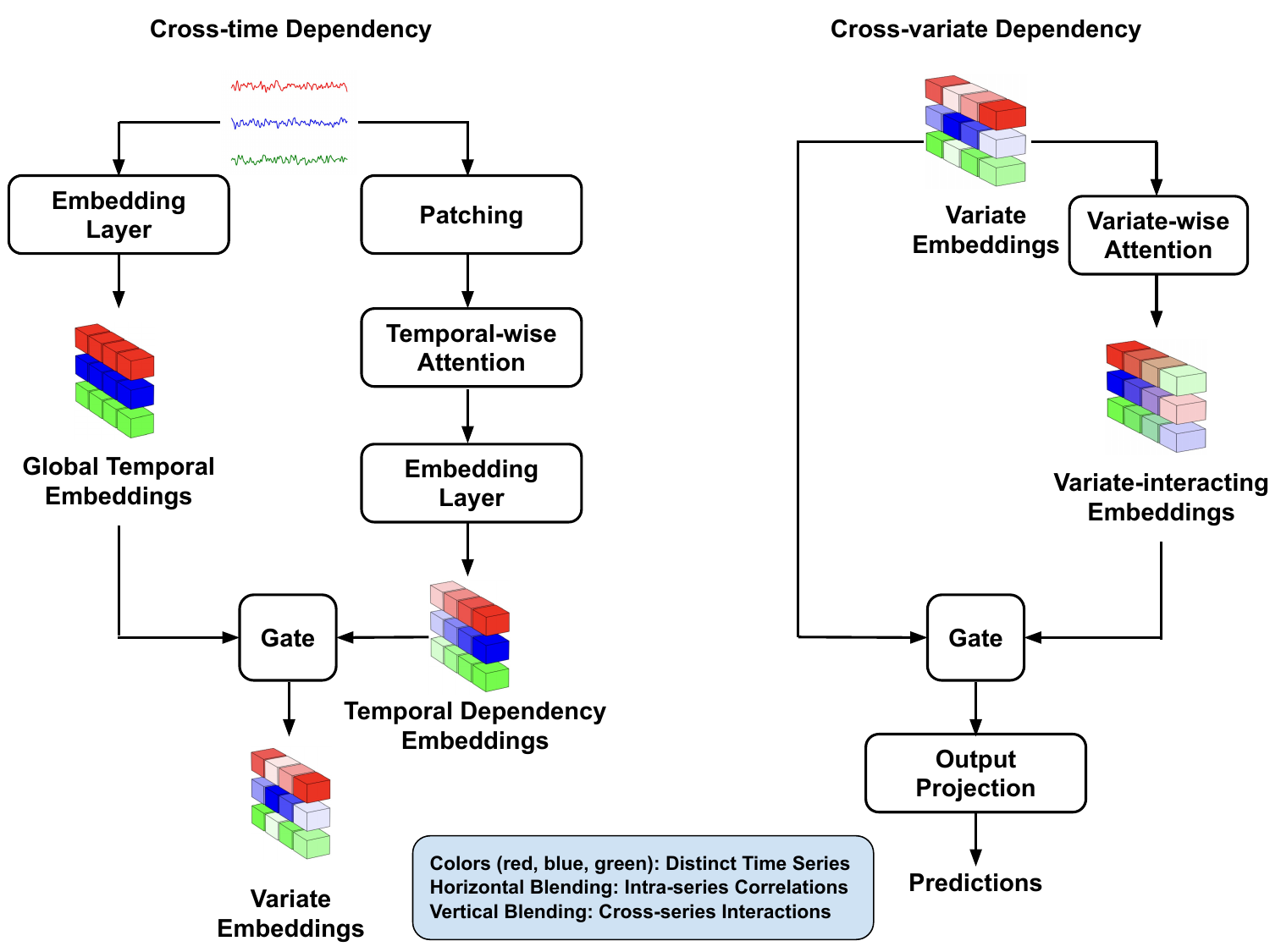}
  \caption{Overall model architecture: The model encodes each variate's series independently through two distinct pathways to obtain variate-wise representations: (1) temporal dependency embeddings that capture cross-time dependencies through patching and temporal-wise attention, and (2) global temporal embeddings that encode global temporal patterns through an MLP. These complementary embeddings are integrated through a gating operation to form variate embeddings, which serve as input for cross-variate dependency modeling. Variate-wise attention is then applied on variate embeddings to model multivariate dependencies, producing variate-interacting embeddings. Finally, a copy of the variate embeddings (without interaction) is combined with the variate-interacting embeddings through gating to regulate cross-variate correlations. The resulting output is then passed through a projection layer to generate predictions.}
  \label{fig:model_architecture}
\end{figure*}

\section{Methodology}
\label{sec:methodology}

\subsection{Problem Definition}
Given multivariate time series data \( \mathbf{X} \in \mathbb{R}^{N \times T} \), the goal is to forecast the next \( F \) steps, \( \mathbf{\hat{Y}} \in \mathbb{R}^{N \times F} \), where \(N\) is the number of variates and \(T\) is the look-back window. The objective is to minimize the MSE between \(\hat{\mathbf{Y}}\) and the ground truth \(\mathbf{Y}\).

\begin{table*}[t]
    \centering
    \caption{Multivariate forecasting results with prediction lengths $T \in \{96, 192, 336, 720\}$ and fixed look-back length $L = 96$. Results are averaged from all prediction lengths. A lower value indicates better performance. Full results are listed in the Appendix~\ref{sec:full_results}.}
    \label{t:multivariate-forecasting}
    \resizebox{\textwidth}{!}{
        \begin{tabular}{c|cc|cc|cc|cc|cc|cc|cc|cc|cc|cc}
            \toprule
            Models & \multicolumn{2}{c}{\textbf{Gateformer}} & \multicolumn{2}{c}{iTransformer} & \multicolumn{2}{c}{PatchTST} & \multicolumn{2}{c}{Crossformer} & \multicolumn{2}{c}{FEDformer} & \multicolumn{2}{c}{Autoformer} & \multicolumn{2}{c}{Stationary} & \multicolumn{2}{c}{TimesNet} & \multicolumn{2}{c}{SCINet} & \multicolumn{2}{c}{DLinear} \\
              & \multicolumn{2}{c}{\textbf{(Ours)}} & \multicolumn{2}{c}{\citeyearpar{iTransformer}} & \multicolumn{2}{c}{\citeyearpar{PatchTST}} & \multicolumn{2}{c}{\citeyearpar{Crossformer}} & \multicolumn{2}{c}{\citeyearpar{fedformer}} & \multicolumn{2}{c}{\citeyearpar{Autoformer}} & \multicolumn{2}{c}{\citeyearpar{Stationary}} & \multicolumn{2}{c}{\citeyearpar{Timesnet}} & \multicolumn{2}{c}{\citeyearpar{SCINet}} & \multicolumn{2}{c}{\citeyearpar{DLinear}} \\
            \cmidrule(lr){2-3} \cmidrule(lr){4-5} \cmidrule(lr){6-7} \cmidrule(lr){8-9} \cmidrule(lr){10-11} \cmidrule(lr){12-13} \cmidrule(lr){14-15}\cmidrule(lr){16-17}\cmidrule(lr){18-19}\cmidrule(lr){20-21}
            Metric & MSE & MAE & MSE & MAE & MSE & MAE & MSE & MAE & MSE & MAE & MSE & MAE & MSE & MAE & MSE & MAE & MSE & MAE & MSE & MAE \\
            \midrule
            ETT(Avg) & \textcolor{red}{\textbf{0.373}} & \textcolor{red}{\textbf{0.394}} & 0.383 & 0.399 &  \textcolor{blue}{\underline{0.381}} & \textcolor{blue}{\underline{0.397}} & 0.685 & 0.578 & 0.408 & 0.428 & 0.465 & 0.459 & 0.471 & 0.464 & 0.391 & 0.404 & 0.689 & 0.597 & 0.442 & 0.444 \\
            \cmidrule(lr){1-21}
            Weather & \textcolor{red}{\textbf{0.253}} & \textcolor{red}{\textbf{0.276}} & \textcolor{blue}{\underline{0.258}} & \textcolor{blue}{\underline{0.278}} & 0.259 & 0.281 & 0.259 & 0.315 & 0.309 & 0.360 & 0.338 & 0.382 & 0.288 & 0.314 & 0.259 & 0.287 & 0.292 & 0.363 & 0.265 & 0.317 \\
            \cmidrule(lr){1-21}
            Electricity & \textcolor{red}{\textbf{0.176}} & \textcolor{red}{\textbf{0.267}} & \textcolor{blue}{\underline{0.178}} & \textcolor{blue}{\underline{0.270}} & 0.205 & 0.290 & 0.244 & 0.334 & 0.214 & 0.327 & 0.227 & 0.338 & 0.193 & 0.296 & 0.192 & 0.295 & 0.268 & 0.365 & 0.212 & 0.300 \\
            \cmidrule(lr){1-21}
            Traffic & \textcolor{red}{\textbf{0.412}} & \textcolor{red}{\textbf{0.276}} & \textcolor{blue}{\underline{0.428}} & \textcolor{blue}{\underline{0.282}} & 0.481 & 0.304 & 0.550 & 0.304 & 0.610 & 0.376 & 0.628 & 0.379 & 0.624 & 0.340 & 0.620 & 0.336 & 0.804 & 0.509 & 0.625 & 0.383 \\
            \cmidrule(lr){1-21}
            Exchange & \textcolor{red}{\textbf{0.326}} &  \textcolor{red}{\textbf{0.386}} & 0.360 & \textcolor{blue}{\underline{0.403}} & 0.367 & 0.404 & 0.940 & 0.707 & 0.519 & 0.429 & 0.613 & 0.539 & 0.461 & 0.454 & 0.416 & 0.443 & 0.750 & 0.626 & \textcolor{blue}{\underline{0.354}} & 0.414 \\
            \cmidrule(lr){1-21}
            Solar-Energy &  \textcolor{red}{\textbf{0.224}} &  \textcolor{red}{\textbf{0.258}} & \textcolor{blue}{\underline{0.233}} & \textcolor{blue}{\underline{0.262}} & 0.270 & 0.307 & 0.641 & 0.639 & 0.291 & 0.381 & 0.885 & 0.711 & 0.261 & 0.381 & 0.301 & 0.319 & 0.282 & 0.375 & 0.330 & 0.401 \\
            \cmidrule(lr){1-21}
            PEMS(Avg) & \textcolor{red}{\textbf{0.070}} & \textcolor{red}{\textbf{0.169}} & \textcolor{blue}{\underline{0.072}} &\textcolor{blue}{\underline{0.173}} & 0.104 & 0.211 & 0.103 & 0.195 & 0.137 & 0.253 & 0.346 & 0.431 & 0.089 & 0.193 & 0.091 & 0.193 & \textcolor{blue}{\underline{0.072}} & 0.174 & 0.133 & 0.252 \\
             \cmidrule(lr){1-21}
            $1^{st}$ Count & 7 & 7 & 0 & 0 & 0 & 0 & 0 & 0 & 0 & 0 & 0 & 0 & 0 & 0 & 0 & 0 & 0 & 0 & 0 & 0 \\
            \bottomrule
        \end{tabular}
    }
\end{table*}

\begin{table*}[t]
\centering
\large
\caption{Ablation study of model components across eight datasets, using prediction length $T = 96$ and look-back window $L = 96$.}
\label{t:ablation of structure}
\resizebox{\textwidth}{!}{
\begin{tabular}{ccccccccccccccccccc}
\toprule
\multirow{2}{*}{Models} & \multicolumn{2}{c}{ETTh1} & \multicolumn{2}{c}{ETTm1} & \multicolumn{2}{c}{ETTh2} & \multicolumn{2}{c}{ETTm2} & \multicolumn{2}{c}{Exchange} & \multicolumn{2}{c}{Electricity} & \multicolumn{2}{c}{Traffic} & \multicolumn{2}{c}{Weather} & \multicolumn{2}{c}{Avg. rank} \\
\cmidrule(lr){2-3} \cmidrule(lr){4-5} \cmidrule(lr){6-7} \cmidrule(lr){8-9} \cmidrule(lr){10-11} \cmidrule(lr){12-13} \cmidrule(lr){14-15} \cmidrule(lr){16-17} \cmidrule(lr){18-19}
& MSE & MAE & MSE & MAE & MSE & MAE & MSE & MAE & MSE & MAE & MSE & MAE & MSE & MAE & MSE & MAE & MSE & MAE \\
\midrule
Gateformer & 0.383 & 0.398 & \textbf{0.320} & \textbf{0.360} & \textbf{0.306} & \textbf{0.351} & \textbf{0.176} & \textbf{0.260} & \textbf{0.081} & \textbf{0.199} & 0.146 & \textbf{0.238} & 0.390 & \textbf{0.261} & \textbf{0.168} & \textbf{0.208} & \textbf{1.375} & \textbf{1.125} \\
\cmidrule(lr){1-19}
w/o Temporal-wise Attn. & 0.393 & 0.404 & 0.327 & 0.361 & 0.325 & 0.360 & 0.179 & 0.261 & 0.084 & 0.203 & 0.147 & 0.239 & 0.395 & 0.266 & 0.176 & 0.217 & 3.25 & 3.000\\
\cmidrule(lr){1-19}
w/o Gate in Variate-wise Attn. & 0.385 & 0.400 & 0.327 & 0.362 & 0.334 & 0.367 & 0.182 & 0.264 & 0.082 & 0.200 & \textbf{0.144} & 0.239 & \textbf{0.387} & \textbf{0.261} & \textbf{0.168} & \textbf{0.208} & 2.250 & 2.375\\
\cmidrule(lr){1-19}
w/o Global Temporal Embeddings & \textbf{0.381} & \textbf{0.397} & 0.323 & 0.362 & 0.343 & 0.367 & 0.179 & 0.262 & 0.084 & 0.200 & 0.146 & 0.241 & 0.392 & \textbf{0.261} & 0.172 & 0.212 & 2.750 & 2.500\\
\bottomrule
\end{tabular}
}
\end{table*}

\subsection{Model Overview}
As shown in Figure~\ref{fig:model_architecture}, our model captures both temporal and cross-variate dependencies.

\textbf{Cross-time Dependency Modeling.}
Each variate is processed independently to capture intra-series patterns. Two parallel paths are used: (1) patch-based temporal attention to capture local dependencies, and (2) a shared MLP to learn global temporal patterns. These are fused via a gating mechanism to form variate embeddings.

\textbf{Cross-variate Dependency Modeling.}
Variate embeddings are passed through a variate-wise attention module to model inter-series relationships. To maintain performance on small datasets, these outputs are gated with original variate embeddings to dynamically control cross-variate influence.

\textbf{Output Projection.}
Final embeddings are passed through a linear layer to produce predictions \(\hat{\mathbf{Y}}\), followed by de-normalization. MSE is used for training.

\subsection{Model Details}

\subsubsection{Cross-time Dependency Modeling}
Each series \( \mathbf{x}^{(i)} \in \mathbb{R}^T \) is normalized with RevIN~\citep{kim2022reversible}, then divided into \(P\) non-overlapping patches. Each patch is projected to a \(d_m\)-dimensional space and passed through self-attention:
\[
\mathbf{O}^{(i)}_{cross-time} = \textsc{Softmax}\left(\frac{\mathbf{Q}^{(i)} \mathbf{K}^{(i)\top}}{\sqrt{d_k}}\right) \mathbf{V}^{(i)}
\]
The attention output is reshaped and passed through an FFN to obtain temporal embeddings \( \mathbf{v}^{(i)}_T \in \mathbb{R}^{d_m} \).

\textbf{Global Temporal Pattern.}
To address the loss of global context from patching, each raw series is also processed by a shared MLP to obtain a global embedding \( \mathbf{v}^{(i)}_G \in \mathbb{R}^{d_m} \).

\subsubsection{Gated Fusion}
We fuse \( \mathbf{v}^{(i)}_T \) and \( \mathbf{v}^{(i)}_G \) using a gate:
\[
Gate = \sigma(\mathbf{v}^{(i)}_T \boldsymbol{W}_{g1} + \mathbf{v}^{(i)}_G \boldsymbol{W}_{g2})
\]
\[
\mathbf{s}^{(i)} = Gate \odot \mathbf{v}^{(i)}_T + (1 - Gate) \odot \mathbf{v}^{(i)}_G
\]
\(\mathbf{s}^{(i)}\) serves as the final variate embedding.

\subsubsection{Cross-variate Dependency Modeling}
Given all variate embeddings \(\mathbf{S} \in \mathbb{R}^{N \times d_m}\), we apply self-attention:
\[
\mathbf{O}_{cross-variate} = \textsc{Softmax}\left(\frac{\mathbf{Q} \mathbf{K}^\top}{\sqrt{d_m}}\right) \mathbf{V}
\]
To ensure robustness across dataset sizes, the output is gated with the original \(\mathbf{S}\) embeddings.

\subsubsection{Output Projection Layer}
We apply a linear head on \(\mathbf{O}_{cross-variate}\) to obtain final predictions \(\hat{\mathbf{Y}} \in \mathbb{R}^{N \times F}\). Predictions are de-normalized using stored mean and standard deviation, and MSE is used as the training loss.

\begin{table*}[t]
\centering
\caption{Full results of performance improvements achieved using our proposed framework.}
\label{tab:full_forecasting_promotion}
\vskip 0.05in
\centering
\begin{threeparttable}
\begin{small}
\renewcommand{\multirowsetup}{\centering}
\setlength{\tabcolsep}{6pt}
\begin{tabular}{c|c|c|cc|cc|cc|cc}
\toprule
\multicolumn{3}{c|}{\multirow{2}{*}{{Models}}} &
\multicolumn{2}{c}{\rotatebox{0}{\scalebox{1.0}{Autoformer}}} &
\multicolumn{2}{c}{\rotatebox{0}{\scalebox{1.0}{Flowformer}}} &
\multicolumn{2}{c}{\rotatebox{0}{\scalebox{1.0}{GPT4TS}}} &
\multicolumn{2}{c}{\rotatebox{0}{\scalebox{1.0}{Moment}}} \\
\multicolumn{3}{c|}{} 
&\multicolumn{2}{c}{\scalebox{1.0}{\citeyearpar{Autoformer}}} 
&\multicolumn{2}{c}{\scalebox{1.0}{\citeyearpar{Flowformer}}} 
&\multicolumn{2}{c}{\scalebox{1.0}{\citeyearpar{GPT4TS}}} 
&\multicolumn{2}{c}{\scalebox{1.0}{\citeyearpar{goswami2024momentfamilyopentimeseries}}}  \\
\cmidrule(lr){4-5} \cmidrule(lr){6-7} \cmidrule(lr){8-9} \cmidrule(lr){10-11}
\multicolumn{3}{c|}{Metric} & \scalebox{1.0}{MSE} & \scalebox{1.0}{MAE} & \scalebox{1.0}{MSE} & \scalebox{1.0}{MAE} & \scalebox{1.0}{MSE} & \scalebox{1.0}{MAE} & \scalebox{1.0}{MSE} & \scalebox{1.0}{MAE} \\
\toprule
\multirow{10}{*}{\scalebox{1.0}{Electricity}} & \multirow{5}{*}{Original} & 96 & 0.201 & 0.317 & 0.215 & 0.320 & 0.197 & 0.290 & 0.204 & 0.296\\
 & & 192 & 0.222 & 0.334 & 0.259 & 0.355 & 0.201 & 0.292 & 0.207 & 0.299\\
 & & 336 & 0.231 & 0.338 & 0.296 & 0.383 & 0.217 & 0.309 & 0.219 & 0.310\\
 & & 720 & 0.254 & 0.361 & 0.296 & 0.380 & 0.253 & 0.339 & 0.256 & 0.341\\
\cmidrule(lr){3-11}
 & & Avg & \scalebox{1.0}{0.227} & \scalebox{1.0}{0.338} & \scalebox{1.0}{0.267} & \scalebox{1.0}{0.359} & \scalebox{1.0}{0.217} & \scalebox{1.0}{0.308} & \scalebox{1.0}{0.221} & \scalebox{1.0}{0.311}\\
\cmidrule(lr){2-11}
 & \multirow{5}{*}{+ \textbf{Our framework}} & 96 & \textbf{0.173} & \textbf{0.259} & \textbf{0.144} & \textbf{0.237} & \textbf{0.150} & \textbf{0.241} & \textbf{0.163} & \textbf{0.254}\\
 & & 192 & \textbf{0.180} & \textbf{0.267} & \textbf{0.160} & \textbf{0.251} & \textbf{0.163} & \textbf{0.254} & \textbf{0.177} & \textbf{0.269}\\
 & & 336 & \textbf{0.198} & \textbf{0.285} & \textbf{0.175} & \textbf{0.269} & \textbf{0.180} & \textbf{0.272} & \textbf{0.195} & \textbf{0.283}\\
 & & 720 & \textbf{0.241} & \textbf{0.320} & \textbf{0.205} & \textbf{0.296} & \textbf{0.214} & \textbf{0.301} & \textbf{0.226} & \textbf{0.311}\\
\cmidrule(lr){3-11}
 & & Avg & \textbf{\scalebox{1.0}{0.198}} & \textbf{\scalebox{1.0}{0.283}} & \textbf{\scalebox{1.0}{0.187}} & \textbf{\scalebox{1.0}{0.274}} & \textbf{\scalebox{1.0}{0.177}} & \textbf{\scalebox{1.0}{0.267}} & \textbf{\scalebox{1.0}{0.190}} & \textbf{\scalebox{1.0}{0.279}}\\
\midrule
\multirow{10}{*}{\scalebox{1.0}{Weather}} & \multirow{5}{*}{Original} & 96 & 0.266 & 0.336 & 0.182 & 0.233 & 0.203 & 0.244 & 0.192 & 0.234\\
 & & 192 & 0.307 & 0.367 & 0.250 & 0.288 & 0.247 & 0.277 & 0.246 & 0.278\\
 & & 336 & 0.359 & 0.395 & 0.309 & 0.329 & 0.297 & 0.311 & 0.287 & 0.305\\
 & & 720 & 0.419 & 0.428 & 0.404 & 0.385 & 0.368 & 0.356 & 0.360 & 0.350\\
\cmidrule(lr){3-11}
 & & Avg & \scalebox{1.0}{0.338} & \scalebox{1.0}{0.382} & \scalebox{1.0}{0.286} & \scalebox{1.0}{0.308} & \scalebox{1.0}{0.279} & \scalebox{1.0}{0.297} & \scalebox{1.0}{0.271} & \scalebox{1.0}{0.292}\\
\cmidrule(lr){2-11}
 & \multirow{5}{*}{+ \textbf{Our framework}} & 96 & \textbf{0.175} & \textbf{0.218} & \textbf{0.171} & \textbf{0.213} & \textbf{0.175} & \textbf{0.214} & \textbf{0.184} & \textbf{0.223}\\
 & & 192 & \textbf{0.219} & \textbf{0.258} & \textbf{0.220} & \textbf{0.258} & \textbf{0.228} & \textbf{0.260} & \textbf{0.225} & \textbf{0.259}\\
 & & 336 & \textbf{0.281} & \textbf{0.300} & \textbf{0.275} & \textbf{0.297} & \textbf{0.282} & \textbf{0.301} & \textbf{0.281} & \textbf{0.301}\\
 & & 720 & \textbf{0.353} & \textbf{0.350} & \textbf{0.353} & \textbf{0.349} & \textbf{0.359} & \textbf{0.352} & \textbf{0.356} & \textbf{0.349} \\
\cmidrule(lr){3-11}
 & & Avg & \textbf{\scalebox{1.0}{0.257}} & \textbf{\scalebox{1.0}{0.281}} & \textbf{\scalebox{1.0}{0.255}} & \textbf{\scalebox{1.0}{0.279}} & \textbf{\scalebox{1.0}{0.261}} & \textbf{\scalebox{1.0}{0.282}} & \textbf{\scalebox{1.0}{0.261}} & \textbf{\scalebox{1.0}{0.283}}\\
\bottomrule
\end{tabular}
\end{small}
\end{threeparttable}
\end{table*}

\section{Experiments}
We evaluate our model on short- and long-term forecasting tasks against nine SOTA methods. Results show strong performance and notable gains when applied to Transformer- and LLM-based forecasters. Ablation studies highlight the contribution of each component.

\textbf{Datasets.} We use 13 real-world datasets, including Traffic, Electricity, Weather, Exchange, four ETT subsets~\citep{Autoformer}, Solar-Energy~\citep{LSTNet}, and four PEMS subsets~\citep{SCINet}. Dataset details are provided in Appendix~\ref{sec:dataset_detail}.

\subsection{Forecasting Results}
\paragraph{Baselines}
We compare our model against nine SOTA methods: Transformer-based (Autoformer, FEDformer, Stationary, Crossformer, PatchTST, iTransformer), linear-based (DLinear), and TCN-based (SCINet, TimesNet).

\paragraph{Experimental Setup}
All models use a look-back window of $L=96$ and are trained for 10 epochs. Prediction lengths are $T \in \{3, 6, 12, 24\}$ for PEMS, and $T \in \{96, 192, 336, 720\}$ for others. Baseline results follow \citet{iTransformer}.
\vspace{-10pt}
\paragraph{Main Results}
Table~\ref{t:multivariate-forecasting} shows that Gateformer consistently outperforms all baselines across 13 datasets. Although iTransformer and Crossformer capture cross-variate dependencies, they fall short due to limitations in modeling temporal dynamics or introducing noise through patch-wise interactions. Our method encodes both temporal and global patterns via gated representations, enabling robust cross-variate modeling. This flexibility allows strong performance across dataset scales. On low-dimensional datasets (e.g., ETT), Gateformer matches or exceeds PatchTST, a strong channel-independent baseline, thanks to its gated control of cross-variate interactions.

\subsection{Ablation Study}
We assess component contributions on eight datasets (ETT, Weather, Traffic, Electricity, Exchange) by removing:  
(1) \textbf{Temporal-wise Attention} (keeps only global patterns),  
(2) \textbf{Gate in Variate-wise Attention}, and  
(3) \textbf{Global Temporal Embeddings} (keeps only local patterns). Table~\ref{t:ablation of structure} shows full Gateformer achieves the best results. Removing temporal attention significantly degrades performance, especially on small datasets, confirming its importance. Removing the variate-wise gate slightly improves performance on large datasets but harms it on smaller ones. Overall, the gated integration of temporal and variate-wise dependencies ensures strong, consistent forecasting performance.

\subsection{Framework Generalizability}
\label{sec:framework_generalizability_main}
While capturing cross-variate dependencies is essential, the quadratic cost of self-attention limits scalability with many variates. Our Transformer-based framework addresses this by easily integrating efficient attention modules from models like Autoformer~\citep{Autoformer} and Flowformer~\citep{Flowformer}, and by extending to LLM-based models like GPT4TS~\citep{GPT4TS} or foundation models such as Moment~\citep{goswami2024momentfamilyopentimeseries}. On Weather and Electricity datasets, our framework improves Autoformer by 19.9\%, Flowformer by 20.7\%, GPT4TS by 10.8\% and Moment by 6.9\%, as shown in Table~\ref{tab:full_forecasting_promotion}. These gains highlight the flexibility and effectiveness of our approach.

\section{Conclusion and Future Work}
We present a model that captures both temporal and cross-variate dependencies using attention and gating mechanisms for improved forecasting. It achieves state-of-the-art results and boosts performance of existing Transformer-based models by up to 20.7\%. Future work includes large-scale pre-training and broader time series applications.

\section*{Software and Data}
Code, dataset details, metrics, and experimental settings are included in Appendix~\ref{sec:experimental_details}.

\bibliography{example_paper}

\begin{thebibliography}{30}
\providecommand{\natexlab}[1]{#1}
\providecommand{\url}[1]{\texttt{#1}}
\expandafter\ifx\csname urlstyle\endcsname\relax
  \providecommand{\doi}[1]{doi: #1}\else
  \providecommand{\doi}{doi: \begingroup \urlstyle{rm}\Url}\fi

\bibitem[Angryk et~al.(2020)Angryk, Martens, Aydin, Kempton, Mahajan, Basodi,
  Ahmadzadeh, Cai, Filali~Boubrahimi, Hamdi, et~al.]{angryk2020multivariate}
Angryk, R.~A., Martens, P.~C., Aydin, B., Kempton, D., Mahajan, S.~S., Basodi,
  S., Ahmadzadeh, A., Cai, X., Filali~Boubrahimi, S., Hamdi, S.~M., et~al.
\newblock Multivariate time series dataset for space weather data analytics.
\newblock \emph{Scientific data}, 7\penalty0 (1):\penalty0 227, 2020.

\bibitem[Chen et~al.(2012)Chen, Gerlach, Lin, and Lee]{chen2012bayesian}
Chen, C.~W., Gerlach, R., Lin, E.~M., and Lee, W.
\newblock Bayesian forecasting for financial risk management, pre and post the
  global financial crisis.
\newblock \emph{Journal of Forecasting}, 31\penalty0 (8):\penalty0 661--687,
  2012.

\bibitem[Das et~al.(2024)Das, Kong, Sen, and Zhou]{TimesFM}
Das, A., Kong, W., Sen, R., and Zhou, Y.
\newblock A decoder-only foundation model for time-series forecasting.
\newblock In \emph{Forty-first International Conference on Machine Learning},
  2024.
\newblock URL \url{https://openreview.net/forum?id=jn2iTJas6h}.

\bibitem[Goswami et~al.(2024)Goswami, Szafer, Choudhry, Cai, Li, and
  Dubrawski]{goswami2024momentfamilyopentimeseries}
Goswami, M., Szafer, K., Choudhry, A., Cai, Y., Li, S., and Dubrawski, A.
\newblock Moment: A family of open time-series foundation models, 2024.
\newblock URL \url{https://arxiv.org/abs/2402.03885}.

\bibitem[Han et~al.(2024)Han, Ye, and Zhan]{han2024capacity}
Han, L., Ye, H.-J., and Zhan, D.-C.
\newblock The capacity and robustness trade-off: Revisiting the channel
  independent strategy for multivariate time series forecasting.
\newblock \emph{IEEE Transactions on Knowledge and Data Engineering}, 2024.

\bibitem[Jin et~al.(2024)Jin, Wang, Ma, Chu, Zhang, Shi, Chen, Liang, Li, Pan,
  and Wen]{TimeLLM}
Jin, M., Wang, S., Ma, L., Chu, Z., Zhang, J.~Y., Shi, X., Chen, P.-Y., Liang,
  Y., Li, Y.-F., Pan, S., and Wen, Q.
\newblock Time-{LLM}: Time series forecasting by reprogramming large language
  models.
\newblock In \emph{The Twelfth International Conference on Learning
  Representations}, 2024.
\newblock URL \url{https://openreview.net/forum?id=Unb5CVPtae}.

\bibitem[Kaushik et~al.(2020)Kaushik, Choudhury, Sheron, Dasgupta, Natarajan,
  Pickett, and Dutt]{kaushik2020ai}
Kaushik, S., Choudhury, A., Sheron, P.~K., Dasgupta, N., Natarajan, S.,
  Pickett, L.~A., and Dutt, V.
\newblock Ai in healthcare: time-series forecasting using statistical, neural,
  and ensemble architectures.
\newblock \emph{Frontiers in big data}, 3:\penalty0 4, 2020.

\bibitem[Kim et~al.(2022)Kim, Kim, Tae, Park, Choi, and
  Choo]{kim2022reversible}
Kim, T., Kim, J., Tae, Y., Park, C., Choi, J.-H., and Choo, J.
\newblock Reversible instance normalization for accurate time-series
  forecasting against distribution shift.
\newblock In \emph{International Conference on Learning Representations}, 2022.
\newblock URL \url{https://openreview.net/forum?id=cGDAkQo1C0p}.

\bibitem[Kitaev et~al.(2020)Kitaev, Kaiser, and Levskaya]{Reformer}
Kitaev, N., Kaiser, L., and Levskaya, A.
\newblock Reformer: The efficient transformer.
\newblock In \emph{International Conference on Learning Representations}, 2020.
\newblock URL \url{https://openreview.net/forum?id=rkgNKkHtvB}.

\bibitem[Lai et~al.(2018)Lai, Chang, Yang, and Liu]{LSTNet}
Lai, G., Chang, W.-C., Yang, Y., and Liu, H.
\newblock Modeling long- and short-term temporal patterns with deep neural
  networks, 2018.
\newblock URL \url{https://arxiv.org/abs/1703.07015}.

\bibitem[Li et~al.(2021)Li, Hui, and Zhang]{Informer}
Li, J., Hui, X., and Zhang, W.
\newblock Informer: Beyond efficient transformer for long sequence time-series
  forecasting.
\newblock \emph{arXiv: 2012.07436}, 2021.

\bibitem[Liu et~al.(2022{\natexlab{a}})Liu, Zeng, Chen, Xu, LAI, Ma, and
  Xu]{SCINet}
Liu, M., Zeng, A., Chen, M., Xu, Z., LAI, Q., Ma, L., and Xu, Q.
\newblock {SCIN}et: Time series modeling and forecasting with sample
  convolution and interaction.
\newblock In Oh, A.~H., Agarwal, A., Belgrave, D., and Cho, K. (eds.),
  \emph{Advances in Neural Information Processing Systems}, 2022{\natexlab{a}}.
\newblock URL \url{https://openreview.net/forum?id=AyajSjTAzmg}.

\bibitem[Liu et~al.(2022{\natexlab{b}})Liu, Yu, Liao, Li, Lin, Liu, and
  Dustdar]{Pyraformer}
Liu, S., Yu, H., Liao, C., Li, J., Lin, W., Liu, A.~X., and Dustdar, S.
\newblock Pyraformer: Low-complexity pyramidal attention for long-range time
  series modeling and forecasting.
\newblock In \emph{International Conference on Learning Representations},
  2022{\natexlab{b}}.
\newblock URL \url{https://openreview.net/forum?id=0EXmFzUn5I}.

\bibitem[Liu et~al.(2024{\natexlab{a}})Liu, Hu, Li, Diao, Liang, Hooi, and
  Zimmermann]{liu2024unitime}
Liu, X., Hu, J., Li, Y., Diao, S., Liang, Y., Hooi, B., and Zimmermann, R.
\newblock Unitime: A language-empowered unified model for cross-domain time
  series forecasting.
\newblock In \emph{The Web Conference 2024}, 2024{\natexlab{a}}.
\newblock URL \url{https://openreview.net/forum?id=P6sKyx2xAB}.

\bibitem[Liu et~al.(2022{\natexlab{c}})Liu, Wu, Wang, and Long]{Stationary}
Liu, Y., Wu, H., Wang, J., and Long, M.
\newblock Non-stationary transformers: Exploring the stationarity in time
  series forecasting.
\newblock \emph{Advances in Neural Information Processing Systems},
  35:\penalty0 9881--9893, 2022{\natexlab{c}}.

\bibitem[Liu et~al.(2024{\natexlab{b}})Liu, Hu, Zhang, Wu, Wang, Ma, and
  Long]{iTransformer}
Liu, Y., Hu, T., Zhang, H., Wu, H., Wang, S., Ma, L., and Long, M.
\newblock itransformer: Inverted transformers are effective for time series
  forecasting.
\newblock In \emph{The Twelfth International Conference on Learning
  Representations}, 2024{\natexlab{b}}.
\newblock URL \url{https://openreview.net/forum?id=JePfAI8fah}.

\bibitem[Liu et~al.(2024{\natexlab{c}})Liu, Zhang, Li, Huang, Wang, and
  Long]{Timer}
Liu, Y., Zhang, H., Li, C., Huang, X., Wang, J., and Long, M.
\newblock Timer: Generative pre-trained transformers are large time series
  models.
\newblock In \emph{Forty-first International Conference on Machine Learning},
  2024{\natexlab{c}}.
\newblock URL \url{https://openreview.net/forum?id=bYRYb7DMNo}.

\bibitem[Lv et~al.(2014)Lv, Duan, Kang, Li, and Wang]{lv2014traffic}
Lv, Y., Duan, Y., Kang, W., Li, Z., and Wang, F.-Y.
\newblock Traffic flow prediction with big data: A deep learning approach.
\newblock \emph{Ieee transactions on intelligent transportation systems},
  16\penalty0 (2):\penalty0 865--873, 2014.

\bibitem[Nie et~al.(2023)Nie, Nguyen, Sinthong, and Kalagnanam]{PatchTST}
Nie, Y., Nguyen, N.~H., Sinthong, P., and Kalagnanam, J.
\newblock A time series is worth 64 words: Long-term forecasting with
  transformers.
\newblock In \emph{The Eleventh International Conference on Learning
  Representations}, 2023.
\newblock URL \url{https://openreview.net/forum?id=Jbdc0vTOcol}.

\bibitem[Paszke et~al.(2019)Paszke, Gross, Massa, Lerer, Bradbury, Chanan,
  Killeen, Lin, Gimelshein, Antiga, et~al.]{Pytorch}
Paszke, A., Gross, S., Massa, F., Lerer, A., Bradbury, J., Chanan, G., Killeen,
  T., Lin, Z., Gimelshein, N., Antiga, L., et~al.
\newblock Pytorch: An imperative style, high-performance deep learning library.
\newblock \emph{Advances in neural information processing systems}, 32, 2019.

\bibitem[Vaswani et~al.(2017)Vaswani, Shazeer, Parmar, Uszkoreit, Jones, Gomez,
  Kaiser, and Polosukhin]{Transformer}
Vaswani, A., Shazeer, N., Parmar, N., Uszkoreit, J., Jones, L., Gomez, A.~N.,
  Kaiser, L.~u., and Polosukhin, I.
\newblock Attention is all you need.
\newblock In \emph{Advances in Neural Information Processing Systems}, 2017.
\newblock URL
  \url{https://proceedings.neurips.cc/paper_files/paper/2017/file/3f5ee243547dee91fbd053c1c4a845aa-Paper.pdf}.

\bibitem[Wen et~al.(2023)Wen, Zhou, Zhang, Chen, Ma, Yan, and
  Sun]{wen2022transformers}
Wen, Q., Zhou, T., Zhang, C., Chen, W., Ma, Z., Yan, J., and Sun, L.
\newblock Transformers in time series: A survey, 2023.
\newblock URL \url{https://arxiv.org/abs/2202.07125}.

\bibitem[Wu et~al.(2021)Wu, Xu, Wang, and Long]{Autoformer}
Wu, H., Xu, J., Wang, J., and Long, M.
\newblock Autoformer: Decomposition transformers with auto-correlation for
  long-term series forecasting.
\newblock In Beygelzimer, A., Dauphin, Y., Liang, P., and Vaughan, J.~W.
  (eds.), \emph{Advances in Neural Information Processing Systems}, 2021.
\newblock URL \url{https://openreview.net/forum?id=J4gRj6d5Qm}.

\bibitem[Wu et~al.(2022)Wu, Wu, Xu, Wang, and Long]{Flowformer}
Wu, H., Wu, J., Xu, J., Wang, J., and Long, M.
\newblock Flowformer: Linearizing transformers with conservation flows.
\newblock \emph{arXiv preprint arXiv:2202.06258}, 2022.

\bibitem[Wu et~al.(2023)Wu, Hu, Liu, Zhou, Wang, and Long]{Timesnet}
Wu, H., Hu, T., Liu, Y., Zhou, H., Wang, J., and Long, M.
\newblock Timesnet: Temporal 2d-variation modeling for general time series
  analysis.
\newblock In \emph{The Eleventh International Conference on Learning
  Representations}, 2023.
\newblock URL \url{https://openreview.net/forum?id=ju_Uqw384Oq}.

\bibitem[Zeng et~al.(2023)Zeng, Chen, Zhang, and Xu]{DLinear}
Zeng, A., Chen, M., Zhang, L., and Xu, Q.
\newblock Are transformers effective for time series forecasting?
\newblock In \emph{Proceedings of the AAAI conference on artificial
  intelligence}, 2023.
\newblock URL
  \url{https://ojs.aaai.org/index.php/AAAI/article/view/26317/26089}.

\bibitem[Zhang \& Yan(2023)Zhang and Yan]{Crossformer}
Zhang, Y. and Yan, J.
\newblock Crossformer: Transformer utilizing cross-dimension dependency for
  multivariate time series forecasting.
\newblock In \emph{The Eleventh International Conference on Learning
  Representations}, 2023.
\newblock URL \url{https://openreview.net/forum?id=vSVLM2j9eie}.

\bibitem[Zhou et~al.(2022)Zhou, Ma, Wen, Wang, Sun, and Jin]{fedformer}
Zhou, T., Ma, Z., Wen, Q., Wang, X., Sun, L., and Jin, R.
\newblock Fedformer: Frequency enhanced decomposed transformer for long-term
  series forecasting, 2022.
\newblock URL \url{https://arxiv.org/abs/2201.12740}.

\bibitem[Zhou et~al.(2023)Zhou, Niu, Wang, Sun, and Jin]{GPT4TS}
Zhou, T., Niu, P., Wang, X., Sun, L., and Jin, R.
\newblock One fits all: Power general time series analysis by pretrained {LM}.
\newblock In \emph{Thirty-seventh Conference on Neural Information Processing
  Systems}, 2023.
\newblock URL \url{https://openreview.net/forum?id=gMS6FVZvmF}.

\bibitem[Zhu et~al.(2023)Zhu, Chen, Xia, Zhou, Niu, Peng, Wang, Liu, Ma, Gu,
  et~al.]{zhu2023energy}
Zhu, Z., Chen, W., Xia, R., Zhou, T., Niu, P., Peng, B., Wang, W., Liu, H., Ma,
  Z., Gu, X., et~al.
\newblock Energy forecasting with robust, flexible, and explainable machine
  learning algorithms.
\newblock \emph{AI Magazine}, 44\penalty0 (4):\penalty0 377--393, 2023.

\end{thebibliography}
\bibliographystyle{icml2025}

\newpage
\appendix
\onecolumn
\section{Related work}
\begin{figure}
  \centering
\includegraphics[width=1.0\linewidth]{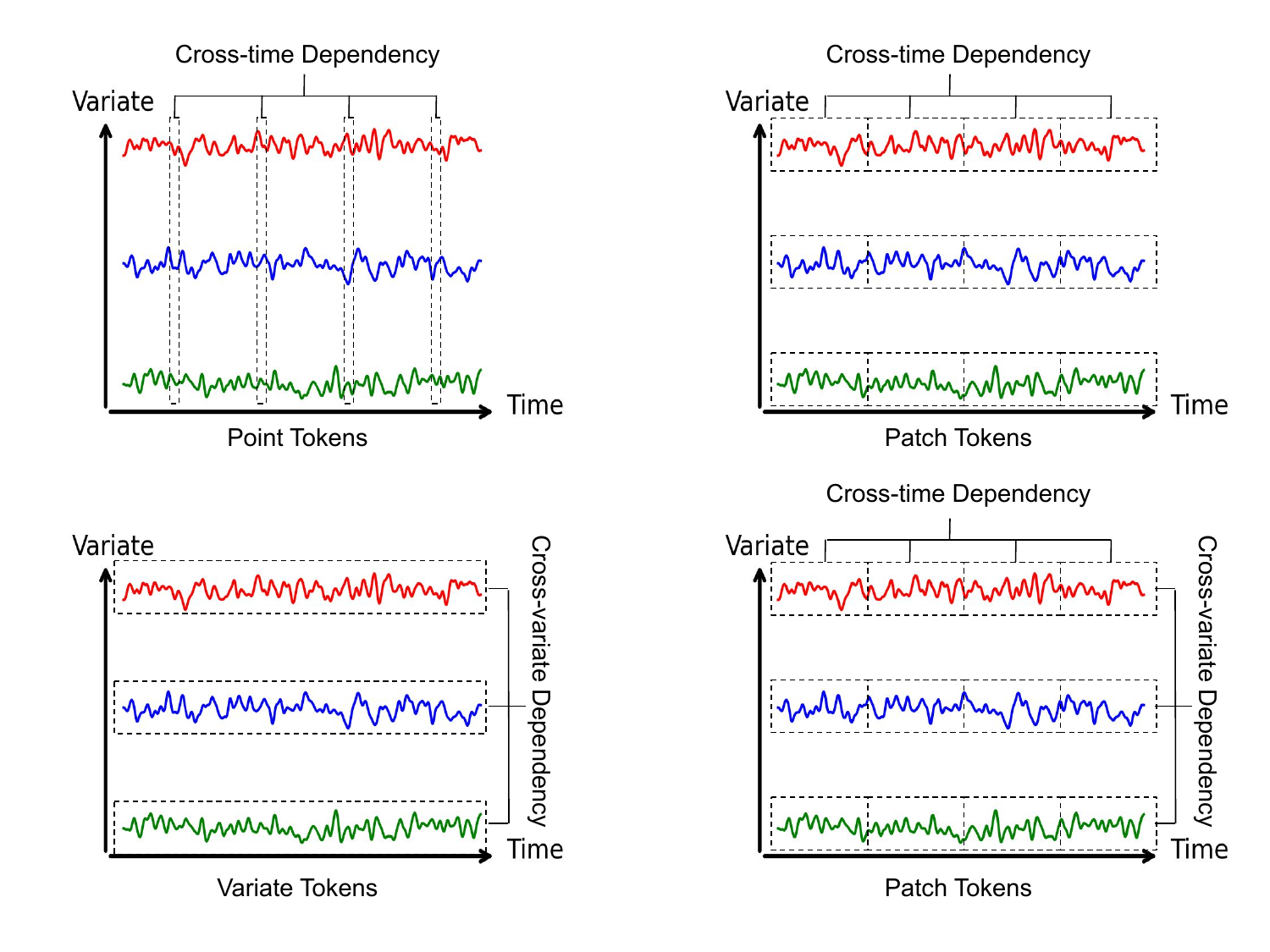}
  \caption{Transformer-based forecasters can be categorized based on their attention granularity (point-wise, patch-wise, and variate-wise) and their approach to modeling cross-time and cross-variate dependencies.}
  \label{fig:related_work}
  \vspace{-10pt}
\end{figure}

As shown in Figure~\ref{fig:related_work}, Transformer-based time series models can be categorized based on their granularity of representations (point-wise, patch-wise, and variate-wise) and their approach to modeling cross-time and cross-variate dependencies.

\textbf{Point-wise Representations with Cross-time Attention}\\
Many prior works embed multiple variates at the same timestamp into point tokens and apply attention mechanisms to capture temporal dependencies among them. Examples include Autoformer \citep{Autoformer}, FEDformer \citep{fedformer}, and Pyraformer \citep{Pyraformer}, which focus on optimizing the quadratic complexity of self-attention. Autoformer detects sub-series similarities with $O(L\log L)$ complexity, FEDformer leverages frequency-domain sparsity with $O(L)$ complexity, and Pyraformer uses pyramidal attention for short- and long-term dependencies with $O(L)$ complexity. However, these models often fail in multivariate forecasting due to the lack of explicit semantic meaning in point tokens and the loss of cross-variate correlations when merging all variables into single temporal tokens.

\textbf{Patch-wise Representations with Cross-time Attention}
PatchTST \citep{PatchTST} addresses the lack of local semantic information through patching. It divides time series into patches to expand the receptive field and applies self-attention to capture cross-time dependencies among patches. PatchTST processes each variate independently, allowing unique attention patterns to be learned for each series. This approach has consistently improved performance on benchmarks and inspired recent large-scale time series models \citep{GPT4TS, TimesFM, TimeLLM, Timer}. However, it focuses primarily on cross-time dependencies while neglecting critical cross-variate interactions.

\textbf{Variate-wise Representations with Cross-variate Attention}
A notable example is iTransformer \citep{iTransformer}, which expands the receptive field as an extreme form of patching. iTransformer encodes each variate's time series into coarse variate-wise representations through linear projection and explicitly models cross-variate correlations among these representations. However, it does not account for cross-time (intra-series) dependencies.

\textbf{Patch-wise Representations with Cross-time and Cross-variate Attention}
Crossformer \citep{Crossformer}, a representative model, segments each variate's series into a sequence of patches and captures both cross-time (intra-series) and cross-variate (inter-series) dependencies among them. However, modeling multivariate correlations at such a granular level can introduce unnecessary noise, potentially resulting in suboptimal forecasting performance.

Unlike previous works, our approach captures both cross-time and cross-variate correlations through variate-wise representations. To enhance the quality of these representations, we integrate gating mechanisms to regulate information flow, allowing the model to focus on the most relevant features for accurate predictions.

\section{Experimental Details}\label{sec:experimental_details}
\subsection{Datasets}\label{sec:dataset_detail}
We assess long-term forecasting performance on ten widely used benchmarks, including the four ETT datasets (ETTh1, ETTh2, ETTm1, ETTm2), Weather, Electricity, Traffic, Exchange (from \citet{Autoformer}), and Solar-Energy (from \citet{LSTNet}). Additionally, we evaluate short-term forecasting on the four PEMS datasets (PEMS03, PEMS04, PEMS07, PEMS08) as used in \citet{SCINet}.
We adopt the same data processing and train-validation-test split protocol as TimesNet~\citep{Timesnet}, ensuring datasets are strictly divided chronologically to prevent data leakage. For forecasting, the lookback series length is fixed at $L = 96$ across all datasets. Prediction lengths are set as follows: $T \in \{96, 192, 336, 720\}$ for ETT, Weather, ECL, Traffic, Solar-Energy, and Exchange; $T \in \{3, 6, 12, 24\}$ for PEMS. Detailed dataset information is provided in Table \ref{tab:dataset}.
\begin{table}[thbp]
  \vspace{0pt}
  \caption{Detailed dataset descriptions. The dimension refers to the number of variates in each dataset and the dataset size is organized in (training, validation, testing).}\label{tab:dataset}
  \vskip 0.05in
  \centering
  \begin{threeparttable}
  \begin{small}
  \renewcommand{\multirowsetup}{\centering}
  \setlength{\tabcolsep}{6.5pt}
  \begin{tabular}{l|c|c|c|c|c}
    \toprule
    Dataset & Dim. & Prediction Length & Dataset Size & Frequency& Information \\
    \toprule
     ETTh1, ETTh2 & 7 & \scalebox{0.8}{\{96, 192, 336, 720\}} & (8545, 2881, 2881) & Hourly & Electricity\\
     \midrule
     ETTm1, ETTm2 & 7 & \scalebox{0.8}{\{96, 192, 336, 720\}} & (34465, 11521, 11521) & 15min & Electricity\\
    \midrule
    Exchange & 8 & \scalebox{0.8}{\{96, 192, 336, 720\}} & (5120, 665, 1422) & Daily & Economy \\
    \midrule
    Weather & 21 & \scalebox{0.8}{\{96, 192, 336, 720\}} & (36792, 5271, 10540) & 10min & Weather\\
    \midrule
    ECL & 321 & \scalebox{0.8}{\{96, 192, 336, 720\}} & (18317, 2633, 5261) & Hourly & Electricity \\
    \midrule
    Traffic & 862 & \scalebox{0.8}{\{96, 192, 336, 720\}} & (12185, 1757, 3509) & Hourly & Transportation \\
    \midrule
    Solar-Energy & 137  & \scalebox{0.8}{\{96, 192, 336, 720\}} & (36601, 5161, 10417) & 10min & Energy \\
    \midrule
    PEMS03 & 358 & \scalebox{0.8}{\{3, 6, 12, 24\}} & (15617, 5135, 5135) & 5min & Transportation\\
    \midrule
    PEMS04 & 307 & \scalebox{0.8}{\{3, 6, 12, 24\}} & (10172, 3375, 3375) & 5min & Transportation\\
    \midrule
    PEMS07 & 883 & \scalebox{0.8}{\{3, 6, 12, 24\}} & (16911, 5622, 5622) & 5min & Transportation\\
    \midrule
    PEMS08 & 170 & \scalebox{0.8}{\{3, 6, 12, 24\}} & (10690, 3548, 3548) & 5min & Transportation\\
    \bottomrule
    \end{tabular}
    \end{small}
  \end{threeparttable}
  \vspace{-5pt}
\end{table}
\subsection{Implementation Details}
All experiments were implemented in PyTorch~\citep{Pytorch} and conducted on a single NVIDIA A100 40GB GPU. We used MSE as the loss function, a batch size of 8, and the Adam optimizer. The initial learning rate was selected from \(\{10^{-4}, 5\times 10^{-4}, 10^{-3}\}\), with training running for a maximum of 10 epochs and early stopping applied if validation loss did not improve within three epochs. The number of encoder blocks was chosen from $L \in \{1,2,3,4\}$, and the hidden state dimension was selected from $\{64,128,256,512,1024\}$. Evaluation metrics included mean square error (MSE) and mean absolute error (MAE), with results averaged over three random seeds. Baseline models were reproduced using the TimesNet~\citep{Timesnet} repository, following configurations from their respective original papers or official code.
\begin{figure}
  \centering
\includegraphics[width=1.0\linewidth]{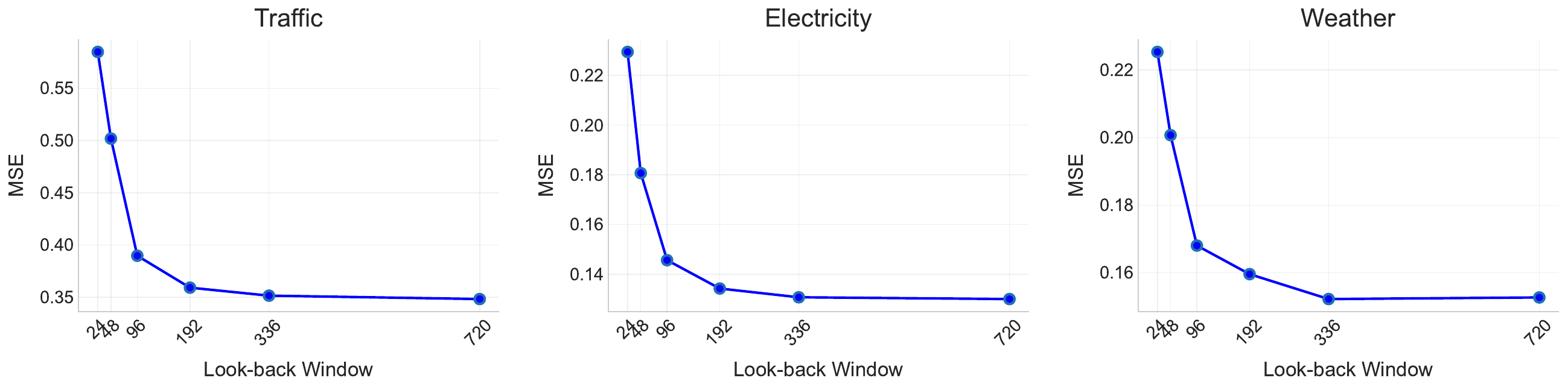}
  \caption{Forecasting performance (MSE) across different look-back windows $L \in \{24,48,96,192,336,720\}$ on Traffic, Electricity, and Weather datasets, with prediction length $T=96$.}
  \label{fig:look_back}
  \vspace{-10pt}
\end{figure}
\section{Varying Look-back Window}
In principle, expanding the look-back window increases the receptive field, which should enhance forecasting accuracy. However, previous studies \citep{DLinear} have shown that most Transformer-based models fail to demonstrate this expected improvement, revealing their limitations in processing extended temporal sequences. Our model, in contrast, effectively utilizes the increased receptive field, achieving lower MSE scores with longer look-back windows, as demonstrated in Figure \ref{fig:look_back}.

\section{Hyperparameter Sensitivity} 
\begin{figure}
  \centering
 \includegraphics[width=1.0\linewidth]{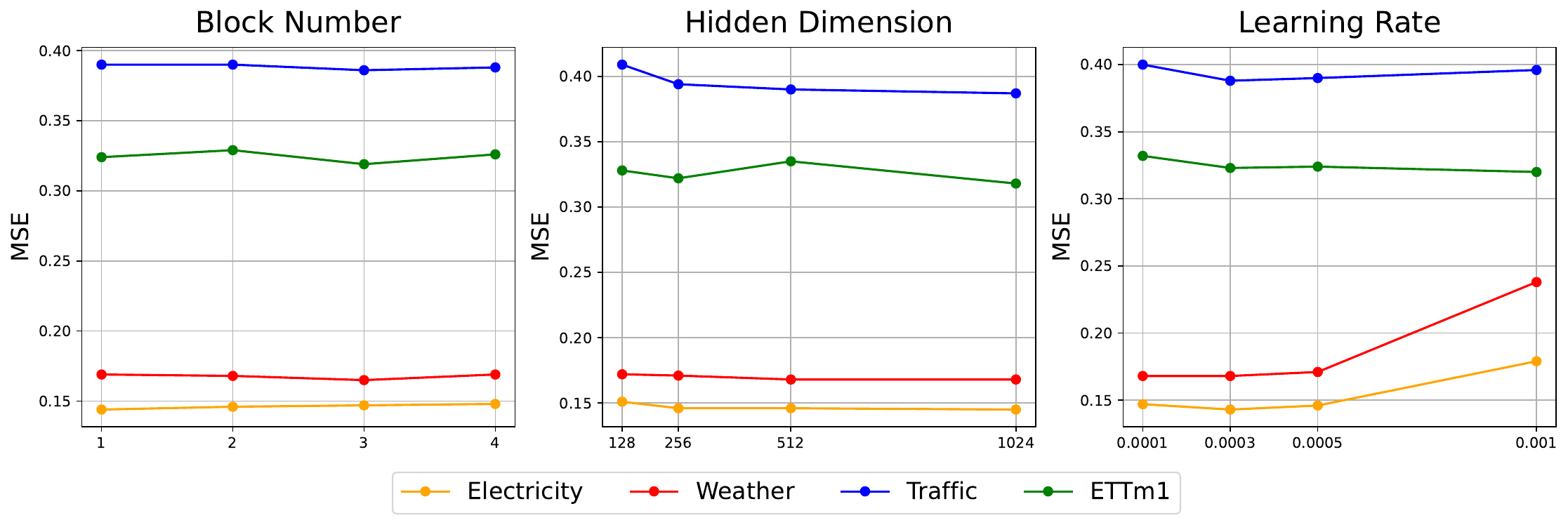}
  \vspace{-15pt}
  \caption{Forecasting results (MSE) varying with the number of Transformer blocks, the model's hidden dimension, and the learning rate. The results were collected with a prediction horizon of $T = 96$ and a look-back window of $L = 96$.}
  \label{fig:hyperparameter}
  \vspace{-15pt}
\end{figure}

To evaluate the sensitivity of our method to hyperparameter settings, we conducted experiments by varying the number of Transformer blocks $L \in \{1,2,3,4\}$, model dimensions $D \in \{128,256,512,1024\}$, and learning rates \(lr \in \{10^{-4}, 3 \times 10^{-4}, 5 \times 10^{-4}, 10^{-3}\}\) on the ETTm1, Weather, Electricity, and Traffic datasets. As shown in Figure \ref{fig:hyperparameter}, the learning rate has the strongest impact on model performance and requires careful tuning. Model performance generally improves with larger hidden dimensions, while remaining relatively stable across different numbers of Transformer blocks.
\section{Framework Generalizability}\label{sec:framework_generalizability}
\begin{figure}
  \centering
\includegraphics[width=1.0\linewidth]{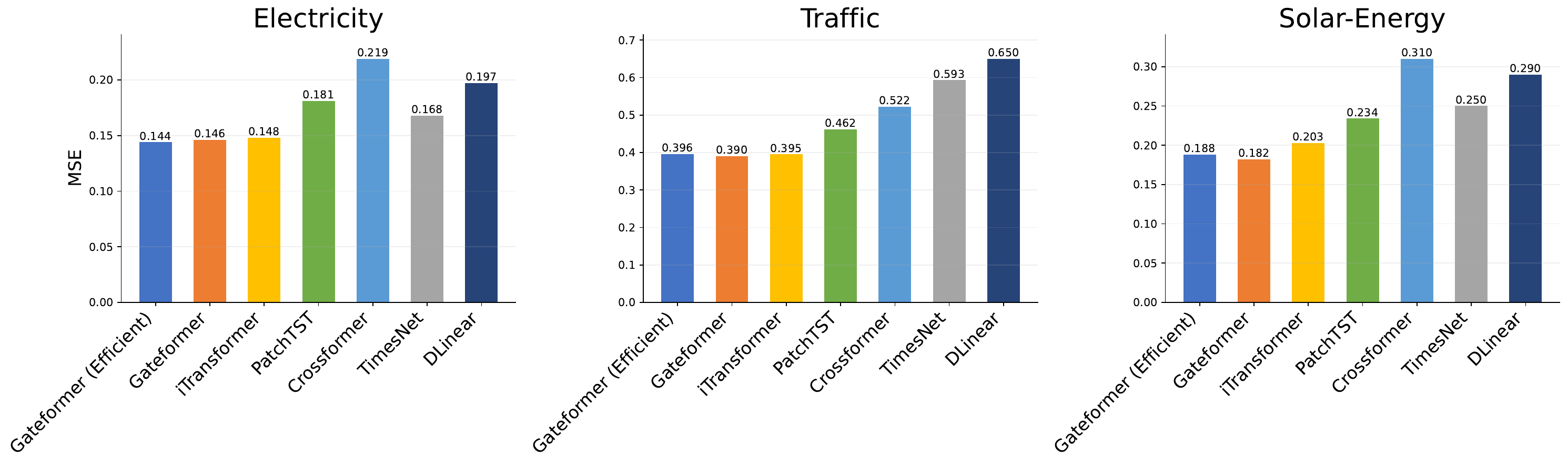}
  \caption{Performance comparison on high-dimensional datasets. Gateformer (Efficient) replaces the conventional quadratic-time attention mechanism with Flowformer's optimized linear-time attention to improve computational efficiency. Prediction horizon $T = 96$; look-back window $L = 96$.}
  \label{fig:comparison_iFlow}
\end{figure}

As discussed in Section~\ref{sec:framework_generalizability_main}, our proposed framework can be seamlessly integrated into other Transformer-based and LLM-based models to consistently improve their performance, with full forecasting results shown in Table~\ref{tab:full_forecasting_promotion}. Since our model explicitly captures cross-variate dependencies, the variate-wise attention becomes a computational bottleneck when training on high-dimensional datasets such as Electricity, Traffic, and Solar-Energy. To mitigate this, we replace the conventional quadratic-time attention mechanism with the linear-time attention proposed in Flowformer \citep{Flowformer} and compare its performance against competitive baselines. These results are provided in Figure~\ref{fig:comparison_iFlow}.

\section{Unified Cross-Dataset Training}
\begin{figure}
  \centering
  \includegraphics[width=0.6\linewidth]{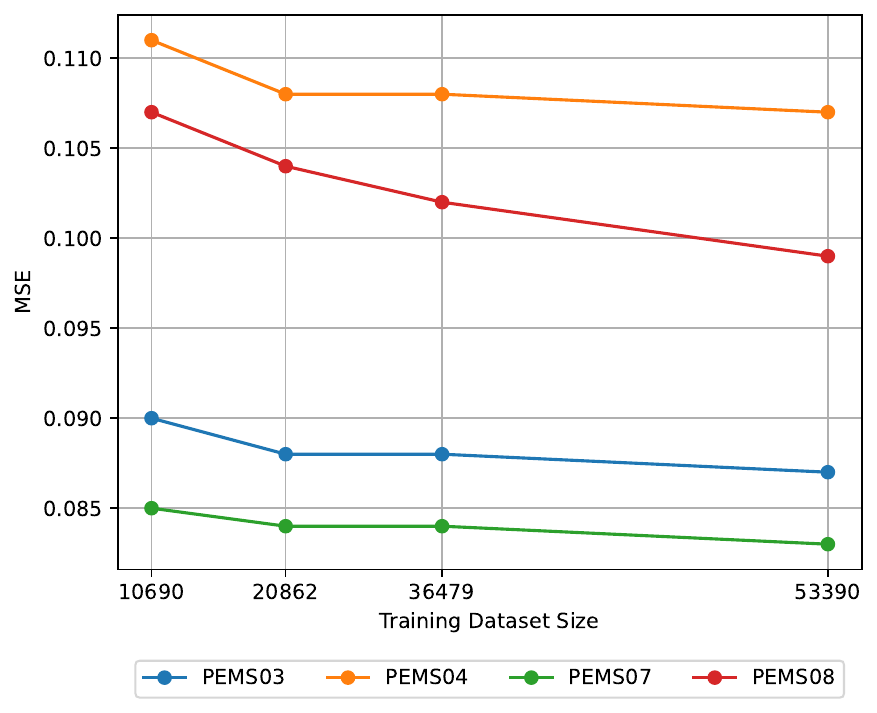}
  \vspace{-15pt}
  \caption{Analysis of unified cross-dataset training: Models are pre-trained on four combinations of PEMS datasets with increasing training set sizes, then fine-tuned on target datasets. Prediction length $T = 24$; look-back window $L = 96$.}
  \label{fig:combined_training}
  \vspace{-15pt}
\end{figure}
We selected the PEMS dataset as our benchmark as it contains four high-dimensional time-series subsets (PEMS03, PEMS04, PEMS07, and PEMS08), each with substantial data that ensures robust evaluation. Our model includes temporal-wise and variate-wise self-attention modules. The temporal-wise attention module captures intra-series dependencies for each series independently. The variate-wise attention module operates on the variate dimension, enabling unified training across datasets with varying numbers of variates. To investigate the benefits of data scaling, we trained our model on four progressively larger combined datasets: (1) PEMS08, (2) PEMS08 + PEMS04, (3) PEMS08 + PEMS04 + PEMS03, and (4) PEMS08 + PEMS04 + PEMS03 + PEMS07. These datasets were combined in ascending order of dimensionality. As shown in Figure \ref{fig:combined_training}, forecasting accuracy improved consistently across all PEMS datasets with larger training set sizes. This highlights our model's ability to learn generalizable and transferable representations through unified training while preventing catastrophic forgetting, a common challenge in combined training \cite{liu2024unitime}. Our results highlight the potential of building large-scale multivariate forecasting models trained across datasets using our framework.

\section{REPRESENTATION LEARNING}
We evaluate the representations learned by our model, focusing on their generalization and transferability across datasets. The PEMS datasets are used as a benchmark, as they comprise four high-dimensional time series subsets, each containing a substantial amount of data to ensure robust and reliable evaluation results. To assess representation transferability, we pre-trained the model on the largest dataset (PEMS07) for 10 epochs, then transferring it to PEMS03, PEMS04, and PEMS08 datasets. As shown in Table \ref{table::transferring performance}, the model’s zero-shot forecasting performance is comparable to most competitive baseline models. A key strength of our method lies in its ability to effectively capture cross-variate correlations, enabling superior generalization and transferability to unseen datasets. Furthermore, pre-training followed by fine-tuning for a few epochs consistently outperforms training solely on downstream datasets, demonstrating the value of transferable representations learned during pre-training.

\begin{table*}[t]
	\centering
    \setlength{\tabcolsep}{4pt}
    \caption{Transfer learning results. Gateformer$^*$ is pre-trained on PEMS07 dataset and then transferred to other datasets. Best results are marked in \textbf{bold}, with prediction lengths $T \in \{3, 6, 12, 24\}$ and look-back window $L = 96$.}
    \vspace{5pt}
	\scalebox{0.9}{
		\begin{tabular}{cc|c|cccc|cccccccccc}
			\cline{2-17}
			&\multicolumn{2}{c|}{\multirow{2}{*}{Models}}& \multicolumn{4}{c|}{Gateformer$^*$}& \multicolumn{10}{c}{Models Trained from Scratch on Target Dataset} \\
			\cline{4-17}
			&\multicolumn{2}{c|}{}& \multicolumn{2}{c}{Zero-shot}& \multicolumn{2}{c|}{Fine-tuning}& \multicolumn{2}{c|}{Gateformer} & \multicolumn{2}{c|}{iTransformer}& \multicolumn{2}{c|}{PatchTST}& \multicolumn{2}{c|}{Crossformer} & \multicolumn{2}{c}{FEDformer}\\
			\cline{2-17}
			&\multicolumn{2}{c|}{Metric}&MSE&MAE&MSE&MAE&MSE&MAE&MSE&MAE&MSE&MAE&MSE&MAE&MSE&MAE \\
			\cline{2-17}
			&\multirow{5}*{\rotatebox{90}{PEMS03}}& 3 & 0.049 & 0.148 & \textbf{0.044} & \textbf{0.140} & 0.045 & 0.142 & 0.047 & 0.147 & 0.055 & 0.166 & 0.051 & 0.150 & 0.109 & 0.231 \\
            &\multicolumn{1}{c|}{}& 6 & 0.064 & 0.167 & \textbf{0.052} & \textbf{0.152} & 0.053 & 0.153 & 0.057 & 0.159 & 0.069 & 0.184 & 0.060 & 0.161 & 0.112 & 0.235 \\
            &\multicolumn{1}{c|}{}& 12 & 0.092 & 0.198 & \textbf{0.065} & \textbf{0.168} & 0.066 & 0.170 & 0.071 & 0.174 & 0.099 & 0.216 & 0.090 & 0.203 & 0.126 & 0.251 \\
            &\multicolumn{1}{c|}{}& 24 & 0.146 & 0.251 & \textbf{0.088} & \textbf{0.197} & 0.092 & 0.201 & 0.093 & 0.201 & 0.142 & 0.259 & 0.121 & 0.240 & 0.149 & 0.275 \\
            &\multicolumn{1}{c|}{}& Avg & 0.088 & 0.191 & \textbf{0.062} & \textbf{0.164} & 0.064 & 0.166 & 0.067 & 0.170 & 0.091 & 0.206 & 0.081 & 0.189 & 0.124 & 0.248 \\
            \cline{2-17}
            &\multirow{5}*{\rotatebox{90}{PEMS04}}& 3 & 0.064 & 0.164 & \textbf{0.060} & \textbf{0.156} & 0.061 & 0.159 & 0.064 & 0.164 & 0.071 & 0.184 & 0.062 & 0.162 & 0.122 & 0.250 \\
            &\multicolumn{1}{c|}{}& 6 & 0.077 & 0.181 & \textbf{0.067} & \textbf{0.166}  & 0.068 & 0.168 & 0.073 & 0.175 & 0.081 & 0.191 & 0.069 & 0.173 & 0.119 & 0.245 \\
            &\multicolumn{1}{c|}{}& 12 & 0.101 & 0.210 & 0.081 & \textbf{0.183} & 0.083 & 0.186 & \textbf{0.078} & \textbf{0.183} & 0.105 & 0.224 & 0.098 & 0.218 & 0.138 & 0.262 \\
            &\multicolumn{1}{c|}{}& 24 & 0.162 & 0.269 & 0.110 & 0.216 & 0.114 & 0.221 & \textbf{0.095} & \textbf{0.205} & 0.153 & 0.275 & 0.131 & 0.256 & 0.177 & 0.293 \\
            &\multicolumn{1}{c|}{}& Avg & 0.101 & 0.206 & 0.080 & 0.180 & 0.081 & 0.184 & \textbf{0.078} & \textbf{0.182} & 0.103 & 0.219 & 0.090 & 0.202 & 0.139 & 0.263 \\
            \cline{2-17}
            &\multirow{5}*{\rotatebox{90}{PEMS08}}& 3 & 0.057 & 0.155 & \textbf{0.051} & \textbf{0.144}  & 0.052 & 0.147 & 0.055 & 0.153 & 0.064 & 0.175 & 0.117 & 0.161 & 0.153 & 0.255 \\
            &\multicolumn{1}{c|}{}& 6 & 0.070 & 0.172 & \textbf{0.059} & \textbf{0.154} & 0.060 & 0.156 & 0.064 & 0.165 & 0.076 & 0.190 & 0.129 & 0.173 & 0.157 & 0.257  \\
            &\multicolumn{1}{c|}{}& 12 & 0.094 & 0.200 & \textbf{0.073} & \textbf{0.170} & 0.075 & 0.172 & 0.079 & 0.182 & 0.168 & 0.232 & 0.165 & 0.214 & 0.173 & 0.273   \\
            &\multicolumn{1}{c|}{}& 24 & 0.157 & 0.258 & \textbf{0.104} & \textbf{0.200} & 0.108 & 0.204 & 0.115 & 0.219 & 0.224 & 0.281 & 0.215 & 0.260 & 0.210 & 0.301 \\
            &\multicolumn{1}{c|}{}& Avg & 0.094 & 0.196 & \textbf{0.072} & \textbf{0.167}  & 0.074 & 0.170 & 0.078 & 0.180 & 0.133 & 0.220 & 0.157 & 0.202 & 0.173 & 0.272 \\
            \cline{2-17}
		\end{tabular}
	}
	\label{table::transferring performance}
\end{table*}

\section{MODEL EFFICIENCY}
\subsection{Memory-Efficient Training Strategy}
\begin{figure}
  \centering
\includegraphics[width=1.0\linewidth]{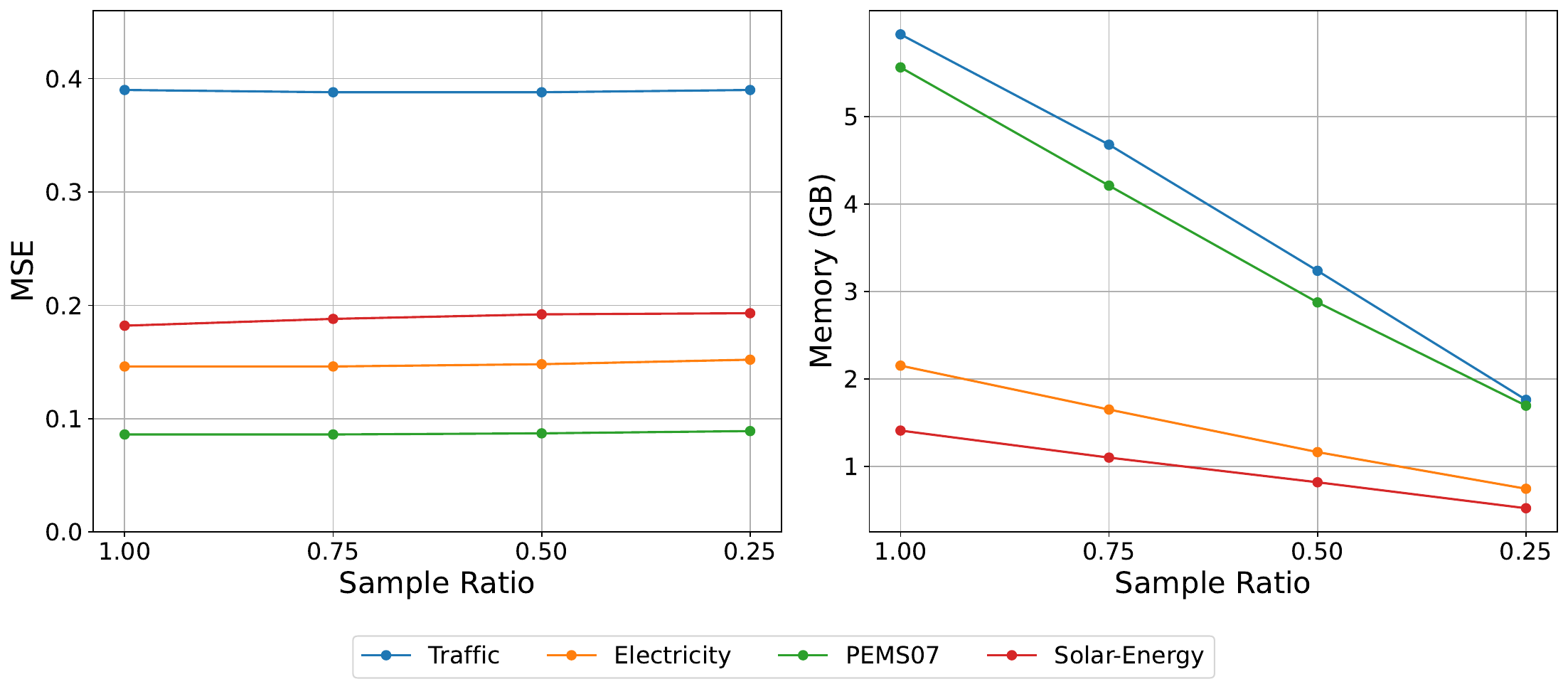}
  \caption{Investigation of the memory-efficient training strategy: The memory footprint (right) is significantly reduced when training on randomly selected variates, while maintaining consistent results (left) across different sampling ratios.
  }
  \label{fig:efficiency1}
  \vspace{-10pt}
\end{figure}

Our model captures multivariate correlations by applying the attention mechanism along the variate dimension. However, the quadratic complexity of the attention mechanism becomes a computational bottleneck when training models on high-dimensional datasets such as Traffic. To mitigate this, we implement a computationally efficient training strategy that randomly samples a subset of variates for each batch, allowing the model to train on only the selected variates. During inference, the model generates forecasts for all variates. This random sampling of variates acts as a form of regularization, enabling the model to learn robust and generalizable patterns. As shown in Figure~\ref{fig:efficiency1}, the forecasting performance remains stable across various sampling ratios, while the memory footprint is significantly reduced.
\subsection{Model Efficiency}
To evaluate the efficiency of our model, we compare the forecasting accuracy, training time, and memory footprint of the following models: Gateformer, Gateformer with the memory-efficient training strategy, iTransformer \citep{iTransformer}, PatchTST \citep{PatchTST}, Crossformer \citep{Crossformer}, TimesNet \citep{Timesnet}, and DLinear \citep{DLinear}. For a fair comparison, all models use the same hidden dimension and batch size on a representative high-dimensional dataset (Traffic).

\begin{figure}
  \centering
\includegraphics[width=1.0\linewidth]{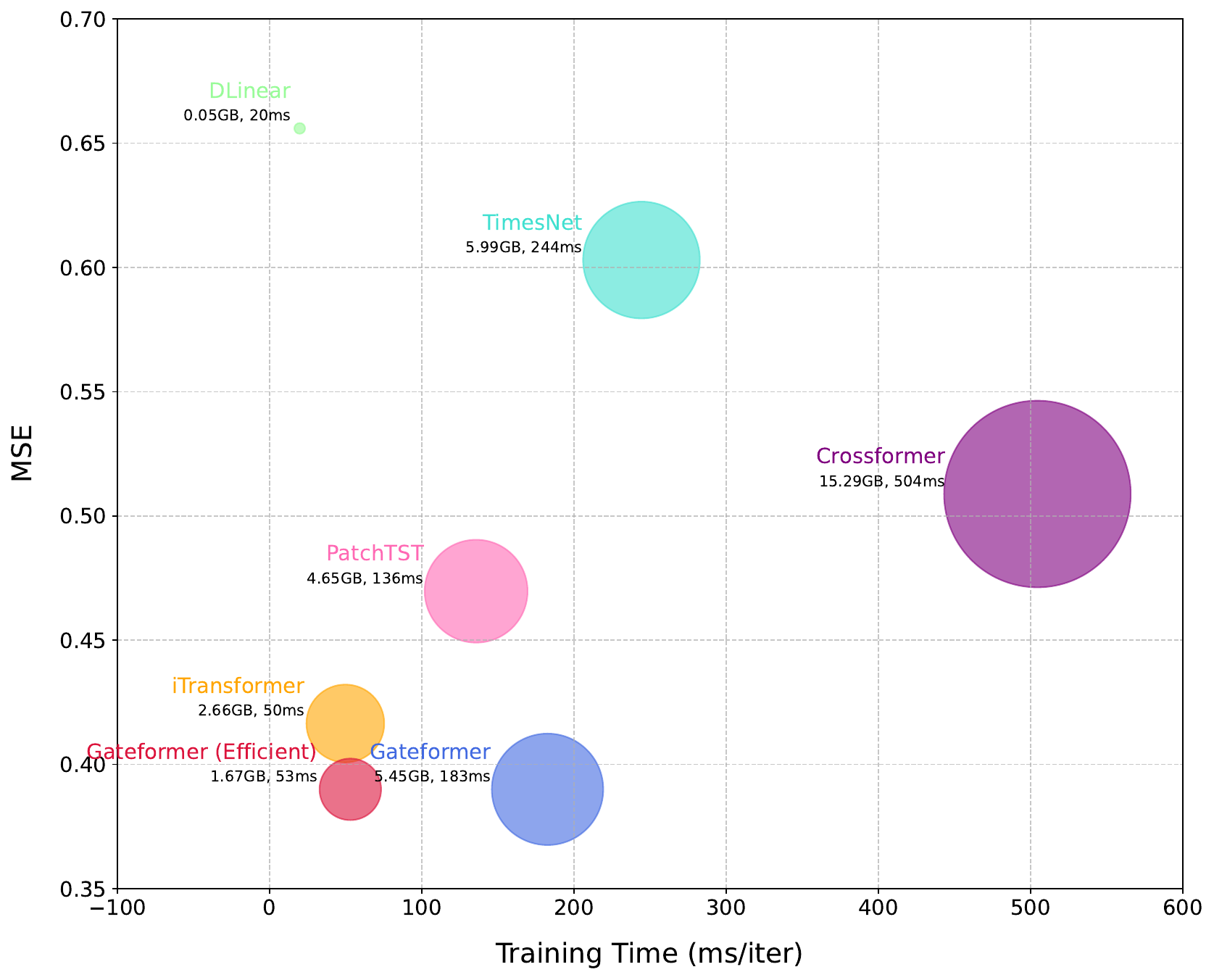}
  \caption{Model efficiency comparison on Traffic dataset.}
  \label{fig:efficiency2}
\end{figure}

As shown in Figure~\ref{fig:efficiency2}, in terms of training speed, the linear model (DLinear) is the fastest. Gateformer, which captures both cross-time and cross-variate dependencies, is slower than iTransformer, which models only multivariate correlations, and PatchTST, which focuses purely on cross-time correlations. However, by adopting an efficient training strategy (training on a randomly sampled 20\% of variates in each batch while forecasting all variates), Gateformer (Efficient) achieves comparable training speed to iTransformer with a lower memory footprint and superior forecasting performance. 

\section{Visualization}
\begin{figure}
  \centering
\includegraphics[width=1.0\linewidth]{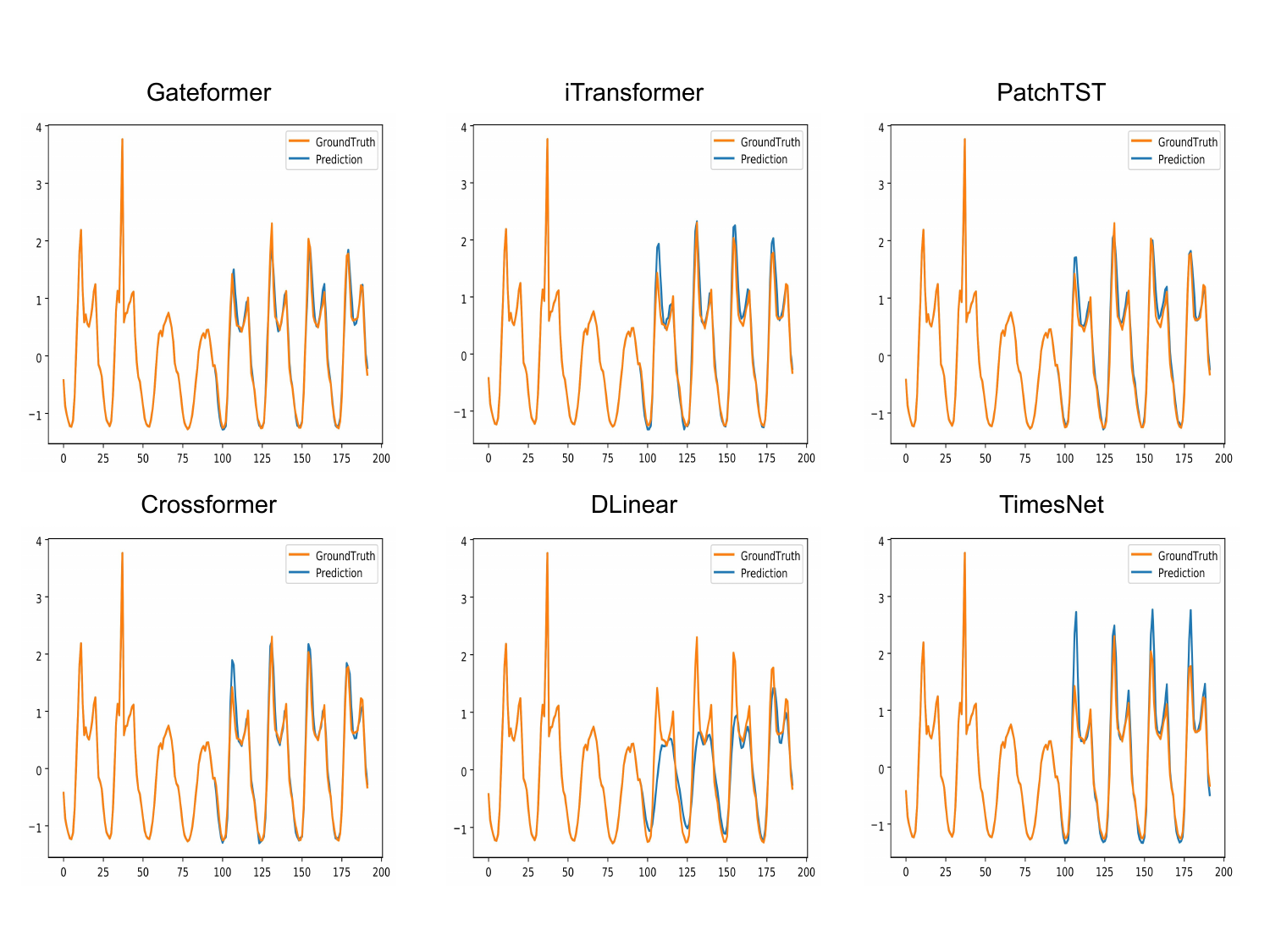}
  \caption{Visualization of 96-step forecasting on Traffic dataset with look-back window $L = 96$.}
  \label{fig:traffic}
\end{figure}

\begin{figure}
  \centering
\includegraphics[width=1.0\linewidth]{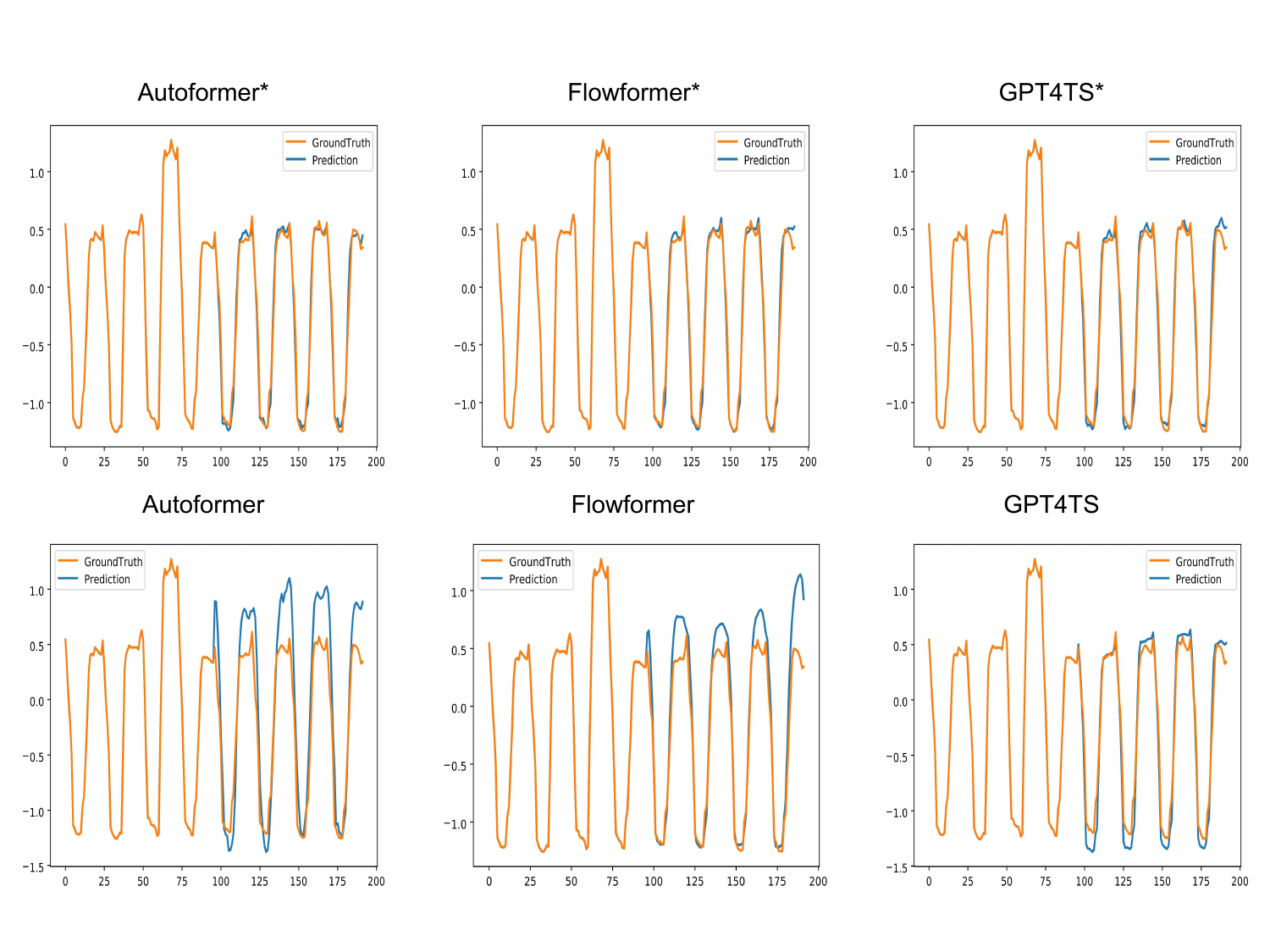}
  \caption{Visualization of 96-step forecasting on Electricity dataset with look-back window $L = 96$. $^*$ denotes models integrated with our framework.}
  \label{fig:ecl}
\end{figure}

We visualize long-term forecasting results of Gateformer against baseline models on the Traffic dataset in Figure~\ref{fig:traffic}, where Gateformer demonstrates superior prediction accuracy. Figure~\ref{fig:ecl} compares the forecasting performance of Transformer-based models with and without integration of our framework on the Electricity dataset. Models integrated with our approach consistently show improved prediction accuracy.

\section{Full Multivariate Forecasting Results}\label{sec:full_results}
We extensively evaluated our model's performance against competitive baselines on well-acknowledged forecasting benchmarks. Table~\ref{tab:full_short_forecasting_results} presents comprehensive short-term forecasting results on the four PEMS subsets. Table~\ref{tab:full_long_forecasting_results} details long-term forecasting results across nine challenging benchmarks.

\begin{table*}
   \vspace{-4mm}
    \caption{Full short-term multivariate forecasting results with prediction lengths $T \in \{3, 6, 12, 24\}$ and input length $L = 96$. The best results are highlighted in \textcolor{red}{\textbf{red}} and the second bests are marked in \textcolor{blue}{\underline{blue}}. }\label{tab:full_short_forecasting_results}
    \vskip 0.05in
    \centering
    \setlength{\tabcolsep}{0.75pt}
    \resizebox{0.90\textwidth}{!}{
    \begin{threeparttable}
    \begin{small}
    \renewcommand{\multirowsetup}{\centering}
    \setlength{\tabcolsep}{1pt}
    \begin{tabular}{c|c|cc|cc|cc|cc|cc|cc|cc|cc|cc|cc}
 \toprule
 \multicolumn{2}{c}{Model} & 
 \multicolumn{2}{c}{\rotatebox{0}{\scalebox{0.8}{\textbf{Gateformer}}}} &\multicolumn{2}{c}{\rotatebox{0}{\scalebox{0.8}{iTransformer}}} &\multicolumn{2}{c}{\rotatebox{0}{\scalebox{0.8}{PatchTST}}} &\multicolumn{2}{c}{\rotatebox{0}{\scalebox{0.8}{Crossformer}}} &\multicolumn{2}{c}{\rotatebox{0}{\scalebox{0.8}{FEDformer}}} &\multicolumn{2}{c}{\rotatebox{0}{\scalebox{0.8}{Autoformer}}} &\multicolumn{2}{c}{\rotatebox{0}{\scalebox{0.8}{Stationary}}} &\multicolumn{2}{c}{\rotatebox{0}{\scalebox{0.8}{TimesNet}}} &\multicolumn{2}{c}{\rotatebox{0}{\scalebox{0.8}{SCINet}}} &\multicolumn{2}{c}{\rotatebox{0}{\scalebox{0.8}{DLinear}}} \\
 
 \multicolumn{2}{c}{ } & \multicolumn{2}{c}{\scalebox{0.76}{(\textbf{Ours})}} & 
 \multicolumn{2}{c}{\scalebox{0.76}{\citeyearpar{iTransformer}}} & 
 \multicolumn{2}{c}{\scalebox{0.76}{\citeyearpar{PatchTST}}} & 
 \multicolumn{2}{c}{\scalebox{0.76}{\citeyearpar{Crossformer}}} & 
 \multicolumn{2}{c}{\scalebox{0.76}{\citeyearpar{fedformer}}} & 
 \multicolumn{2}{c}{\scalebox{0.76}{\citeyearpar{Autoformer}}} & 
 \multicolumn{2}{c}{\scalebox{0.76}{\citeyearpar{Stationary}}} & 
 \multicolumn{2}{c}{\scalebox{0.76}{\citeyearpar{Timesnet}}} & 
 \multicolumn{2}{c}{\scalebox{0.76}{\citeyearpar{SCINet}}} &
 \multicolumn{2}{c}{\scalebox{0.76}{\citeyearpar{DLinear}}} \\
 
 \cmidrule(lr){3-4}\cmidrule(lr){5-6}\cmidrule(lr){7-8}\cmidrule(lr){9-10}\cmidrule(lr){11-12}\cmidrule(lr){13-14}\cmidrule(lr){15-16}\cmidrule(lr){17-18}\cmidrule(lr){19-20}\cmidrule(lr){21-22}
 \multicolumn{2}{c}{Metric} & \scalebox{0.78}{MSE} & \scalebox{0.78}{MAE} & \scalebox{0.78}{MSE} & \scalebox{0.78}{MAE} & \scalebox{0.78}{MSE} & \scalebox{0.78}{MAE} & \scalebox{0.78}{MSE} & \scalebox{0.78}{MAE} & \scalebox{0.78}{MSE} & \scalebox{0.78}{MAE} & \scalebox{0.78}{MSE} & \scalebox{0.78}{MAE} & \scalebox{0.78}{MSE} & \scalebox{0.78}{MAE} & \scalebox{0.78}{MSE} & \scalebox{0.78}{MAE} & \scalebox{0.78}{MSE} & \scalebox{0.78}{MAE} & \scalebox{0.78}{MSE} & \scalebox{0.78}{MAE} \\
 \toprule
 
 \multirow{5}{*}{\rotatebox{90}{\scalebox{0.95}{PEMS03}}} 
& \scalebox{0.78}{3} & \textcolor{red}{\textbf{\scalebox{0.78}{0.045}}} & \textcolor{red}{\textbf{\scalebox{0.78}{0.142}}} & \textcolor{blue}{\underline{\scalebox{0.78}{0.047}}} & \textcolor{blue}{\underline{\scalebox{0.78}{0.147}}} & \scalebox{0.78}{0.055} & \scalebox{0.78}{0.166} & \scalebox{0.78}{0.051} & \scalebox{0.78}{0.150} & \scalebox{0.78}{0.109} & \scalebox{0.78}{0.231} & \scalebox{0.78}{0.285} & \scalebox{0.78}{0.394} & \scalebox{0.78}{0.063} & \scalebox{0.78}{0.168} & \scalebox{0.78}{0.068} & \scalebox{0.78}{0.173} & \scalebox{0.78}{0.048} & \textcolor{blue}{\underline{\scalebox{0.78}{0.147}}} & \scalebox{0.78}{0.069} & \scalebox{0.78}{0.183} \\
& \scalebox{0.78}{6} & \textcolor{red}{\textbf{\scalebox{0.78}{0.053}}} & \textcolor{red}{\textbf{\scalebox{0.78}{0.153}}} & \scalebox{0.78}{0.057} & \scalebox{0.78}{0.159} & \scalebox{0.78}{0.069} & \scalebox{0.78}{0.184} & \scalebox{0.78}{0.060} & \scalebox{0.78}{0.161} & \scalebox{0.78}{0.112} & \scalebox{0.78}{0.235} & \scalebox{0.78}{0.254} & \scalebox{0.78}{0.363} & \scalebox{0.78}{0.070} & \scalebox{0.78}{0.176} & \scalebox{0.78}{0.074} & \scalebox{0.78}{0.179} & \textcolor{blue}{\underline{\scalebox{0.78}{0.056}}} & \textcolor{blue}{\underline{\scalebox{0.78}{0.158}}} & \scalebox{0.78}{0.086} & \scalebox{0.78}{0.204} \\
& \scalebox{0.78}{12} & \textcolor{red}{\textbf{\scalebox{0.78}{0.066}}} & \textcolor{red}{\textbf{\scalebox{0.78}{0.170}}} & \textcolor{blue}{\underline{\scalebox{0.78}{0.071}}} & \scalebox{0.78}{0.174} & \scalebox{0.78}{0.099} & \scalebox{0.78}{0.216} & \scalebox{0.78}{0.090} & \scalebox{0.78}{0.203} & \scalebox{0.78}{0.126} & \scalebox{0.78}{0.251} & \scalebox{0.78}{0.272} & \scalebox{0.78}{0.385} & \scalebox{0.78}{0.081} & \scalebox{0.78}{0.188} & \scalebox{0.78}{0.085} & \scalebox{0.78}{0.192} & \textcolor{red}{\textbf{\scalebox{0.78}{0.066}}} & \textcolor{blue}{\underline{\scalebox{0.78}{0.172}}} & \scalebox{0.78}{0.122} & \scalebox{0.78}{0.243} \\
& \scalebox{0.78}{24} & \textcolor{blue}{\underline{\scalebox{0.78}{0.092}}} & \textcolor{blue}{\underline{\scalebox{0.78}{0.201}}} & \scalebox{0.78}{0.093} & \textcolor{blue}{\underline{\scalebox{0.78}{0.201}}} & \scalebox{0.78}{0.142} & \scalebox{0.78}{0.259} & \scalebox{0.78}{0.121} & \scalebox{0.78}{0.240} & \scalebox{0.78}{0.149} & \scalebox{0.78}{0.275} & \scalebox{0.78}{0.334} & \scalebox{0.78}{0.440} & \scalebox{0.78}{0.105} & \scalebox{0.78}{0.214} & \scalebox{0.78}{0.118} & \scalebox{0.78}{0.223} & \textcolor{red}{\textbf{\scalebox{0.78}{0.085}}} & \textcolor{red}{\textbf{\scalebox{0.78}{0.198}}} & \scalebox{0.78}{0.201} & \scalebox{0.78}{0.317} \\ 
\cmidrule(lr){2-22}
& \scalebox{0.78}{Avg} & \textcolor{red}{\textbf{\scalebox{0.78}{0.064}}} & \textcolor{red}{\textbf{\scalebox{0.78}{0.166}}} & \textcolor{blue}{\underline{\scalebox{0.78}{0.067}}} & \scalebox{0.78}{0.170} & \scalebox{0.78}{0.091} & \scalebox{0.78}{0.206} & \scalebox{0.78}{0.081} & \scalebox{0.78}{0.189} & \scalebox{0.78}{0.124} & \scalebox{0.78}{0.248} & \scalebox{0.78}{0.286} & \scalebox{0.78}{0.396} & \scalebox{0.78}{0.080} & \scalebox{0.78}{0.187} & \scalebox{0.78}{0.086} & \scalebox{0.78}{0.192} & \textcolor{red}{\textbf{\scalebox{0.78}{0.064}}} & \textcolor{blue}{\underline{\scalebox{0.78}{0.169}}} & \scalebox{0.78}{0.120} & \scalebox{0.78}{0.237}\\ \midrule
\multirow{5}{*}{\rotatebox{90}{\scalebox{0.95}{PEMS04}}} 
& \scalebox{0.78}{3} & \textcolor{blue}{\underline{\scalebox{0.78}{0.061}}} & \textcolor{red}{\textbf{\scalebox{0.78}{0.159}}} & \scalebox{0.78}{0.064} & \scalebox{0.78}{0.164} & \scalebox{0.78}{0.071} & \scalebox{0.78}{0.184} & \scalebox{0.78}{0.062} & \textcolor{blue}{\underline{\scalebox{0.78}{0.162}}} & \scalebox{0.78}{0.122} & \scalebox{0.78}{0.250} & \scalebox{0.78}{0.305} & \scalebox{0.78}{0.414} & \scalebox{0.78}{0.076} & \scalebox{0.78}{0.181} & \scalebox{0.78}{0.075} & \scalebox{0.78}{0.179} & \textcolor{red}{\textbf{\scalebox{0.78}{0.060}}} & \textcolor{red}{\textbf{\scalebox{0.78}{0.159}}} & \scalebox{0.78}{0.096} & \scalebox{0.78}{0.218}\\
& \scalebox{0.78}{6} & \textcolor{blue}{\underline{\scalebox{0.78}{0.068}}} & \textcolor{red}{\textbf{\scalebox{0.78}{0.168}}} & \scalebox{0.78}{0.073} & \scalebox{0.78}{0.175} & \scalebox{0.78}{0.081} & \scalebox{0.78}{0.191} & \scalebox{0.78}{0.069} & \scalebox{0.78}{0.173} & \scalebox{0.78}{0.119} & \scalebox{0.78}{0.245} & \scalebox{0.78}{0.361} & \scalebox{0.78}{0.449} & \scalebox{0.78}{0.080} & \scalebox{0.78}{0.187} & \scalebox{0.78}{0.079} & \scalebox{0.78}{0.183} & \textcolor{red}{\textbf{\scalebox{0.78}{0.067}}} & \textcolor{blue}{\underline{\scalebox{0.78}{0.169}}} & \scalebox{0.78}{0.113} & \scalebox{0.78}{0.236}\\
& \scalebox{0.78}{12} & \scalebox{0.78}{0.083} & \scalebox{0.78}{0.186} & \textcolor{blue}{\underline{\scalebox{0.78}{0.078}}} & \textcolor{blue}{\underline{\scalebox{0.78}{0.183}}} & \scalebox{0.78}{0.105} & \scalebox{0.78}{0.224} & \scalebox{0.78}{0.098} & \scalebox{0.78}{0.218} & \scalebox{0.78}{0.138} & \scalebox{0.78}{0.262} & \scalebox{0.78}{0.424} & \scalebox{0.78}{0.491} & \scalebox{0.78}{0.088} & \scalebox{0.78}{0.196} & \scalebox{0.78}{0.087} & \scalebox{0.78}{0.195} & \textcolor{red}{\textbf{\scalebox{0.78}{0.073}}} & \textcolor{red}{\textbf{\scalebox{0.78}{0.177}}} & \scalebox{0.78}{0.148} & \scalebox{0.78}{0.272} \\
& \scalebox{0.78}{24} & \scalebox{0.78}{0.114} & \scalebox{0.78}{0.221} & \textcolor{blue}{\underline{\scalebox{0.78}{0.095}}} & \textcolor{blue}{\underline{\scalebox{0.78}{0.205}}} & \scalebox{0.78}{0.153} & \scalebox{0.78}{0.275} & \scalebox{0.78}{0.131} & \scalebox{0.78}{0.256} & \scalebox{0.78}{0.177} & \scalebox{0.78}{0.293} & \scalebox{0.78}{0.459} & \scalebox{0.78}{0.509} & \scalebox{0.78}{0.104} & \scalebox{0.78}{0.216} & \scalebox{0.78}{0.103} & \scalebox{0.78}{0.215} & \textcolor{red}{\textbf{\scalebox{0.78}{0.084}}} & \textcolor{red}{\textbf{\scalebox{0.78}{0.193}}} & \scalebox{0.78}{0.224} & \scalebox{0.78}{0.340}\\ 
\cmidrule(lr){2-22}
& \scalebox{0.78}{Avg} & \scalebox{0.78}{0.081} & \scalebox{0.78}{0.184} & \textcolor{blue}{\underline{\scalebox{0.78}{0.078}}} & \textcolor{blue}{\underline{\scalebox{0.78}{0.182}}} & \scalebox{0.78}{0.103} & \scalebox{0.78}{0.219} & \scalebox{0.78}{0.090} & \scalebox{0.78}{0.202} & \scalebox{0.78}{0.139} & \scalebox{0.78}{0.263} & \scalebox{0.78}{0.387} & \scalebox{0.78}{0.466} & \scalebox{0.78}{0.087} & \scalebox{0.78}{0.195} & \scalebox{0.78}{0.086} & \scalebox{0.78}{0.193} & \textcolor{red}{\textbf{\scalebox{0.78}{0.071}}} & \textcolor{red}{\textbf{\scalebox{0.78}{0.175}}} & \scalebox{0.78}{0.145} & \scalebox{0.78}{0.267}\\  \midrule
\multirow{5}{*}{\rotatebox{90}{\scalebox{0.95}{PEMS07}}} 
& \scalebox{0.78}{3} & \textcolor{red}{\textbf{\scalebox{0.78}{0.042}}} & \textcolor{red}{\textbf{\scalebox{0.78}{0.130}}} & \scalebox{0.78}{0.046} & \scalebox{0.78}{0.139} & \scalebox{0.78}{0.052} & \scalebox{0.78}{0.158} & \scalebox{0.78}{0.050} & \scalebox{0.78}{0.141} & \scalebox{0.78}{0.102} & \scalebox{0.78}{0.218} & \scalebox{0.78}{0.201} & \scalebox{0.78}{0.326} & \scalebox{0.78}{0.069} & \scalebox{0.78}{0.169} & \scalebox{0.78}{0.068} & \scalebox{0.78}{0.165} & \textcolor{blue}{\underline{\scalebox{0.78}{0.043}}} & \textcolor{blue}{\underline{\scalebox{0.78}{0.132}}} & \scalebox{0.78}{0.061} & \scalebox{0.78}{0.172}\\
& \scalebox{0.78}{6} & \textcolor{red}{\textbf{\scalebox{0.78}{0.049}}} & \textcolor{red}{\textbf{\scalebox{0.78}{0.141}}} & \scalebox{0.78}{0.054} & \scalebox{0.78}{0.150} & \scalebox{0.78}{0.061} & \scalebox{0.78}{0.169} & \scalebox{0.78}{0.057} & \scalebox{0.78}{0.152} & \scalebox{0.78}{0.104} & \scalebox{0.78}{0.219} & \scalebox{0.78}{0.253} & \scalebox{0.78}{0.373} & \scalebox{0.78}{0.074} & \scalebox{0.78}{0.175} & \scalebox{0.78}{0.073} & \scalebox{0.78}{0.171} & \textcolor{blue}{\underline{\scalebox{0.78}{0.050}}} & \textcolor{blue}{\underline{\scalebox{0.78}{0.144}}} & \scalebox{0.78}{0.078} & \scalebox{0.78}{0.196}\\
& \scalebox{0.78}{12} & \textcolor{red}{\textbf{\scalebox{0.78}{0.061}}} & \textcolor{red}{\textbf{\scalebox{0.78}{0.156}}} & \textcolor{blue}{\underline{\scalebox{0.78}{0.067}}} & \textcolor{blue}{\underline{\scalebox{0.78}{0.165}}} & \scalebox{0.78}{0.095} & \scalebox{0.78}{0.207} & \scalebox{0.78}{0.094} & \scalebox{0.78}{0.200} & \scalebox{0.78}{0.109} & \scalebox{0.78}{0.225} & \scalebox{0.78}{0.199} & \scalebox{0.78}{0.336} & \scalebox{0.78}{0.083} & \scalebox{0.78}{0.185} & \scalebox{0.78}{0.082} & \scalebox{0.78}{0.181} & \scalebox{0.78}{0.068} & \scalebox{0.78}{0.171} & \scalebox{0.78}{0.115} & \scalebox{0.78}{0.242}\\
& \scalebox{0.78}{24} & \textcolor{red}{\textbf{\scalebox{0.78}{0.086}}} & \textcolor{red}{\textbf{\scalebox{0.78}{0.187}}} & \textcolor{blue}{\underline{\scalebox{0.78}{0.088}}} & \textcolor{blue}{\underline{\scalebox{0.78}{0.190}}} & \scalebox{0.78}{0.150} & \scalebox{0.78}{0.262} & \scalebox{0.78}{0.139} & \scalebox{0.78}{0.247} & \scalebox{0.78}{0.125} & \scalebox{0.78}{0.244} & \scalebox{0.78}{0.323} & \scalebox{0.78}{0.420} & \scalebox{0.78}{0.102} & \scalebox{0.78}{0.207} & \scalebox{0.78}{0.101} & \scalebox{0.78}{0.204} & \scalebox{0.78}{0.119} & \scalebox{0.78}{0.225} & \scalebox{0.78}{0.210} & \scalebox{0.78}{0.329}\\ 
\cmidrule(lr){2-22}
& \scalebox{0.78}{Avg} & \textcolor{red}{\textbf{\scalebox{0.78}{0.060}}} & \textcolor{red}{\textbf{\scalebox{0.78}{0.154}}} & \textcolor{blue}{\underline{\scalebox{0.78}{0.064}}} & \textcolor{blue}{\underline{\scalebox{0.78}{0.161}}} & \scalebox{0.78}{0.090} & \scalebox{0.78}{0.199} & \scalebox{0.78}{0.085} & \scalebox{0.78}{0.185} & \scalebox{0.78}{0.110} & \scalebox{0.78}{0.227} & \scalebox{0.78}{0.244} & \scalebox{0.78}{0.364} & \scalebox{0.78}{0.082} & \scalebox{0.78}{0.184} & \scalebox{0.78}{0.081} & \scalebox{0.78}{0.180} & \scalebox{0.78}{0.070} & \scalebox{0.78}{0.168} & \scalebox{0.78}{0.116} & \scalebox{0.78}{0.235}\\ \midrule
\multirow{5}{*}{\rotatebox{90}{\scalebox{0.95}{PEMS08}}} 
& \scalebox{0.78}{3} & \textcolor{red}{\textbf{\scalebox{0.78}{0.052}}} & \textcolor{red}{\textbf{\scalebox{0.78}{0.147}}} & \textcolor{blue}{\underline{\scalebox{0.78}{0.055}}} & \textcolor{blue}{\underline{\scalebox{0.78}{0.153}}} & \scalebox{0.78}{0.064} & \scalebox{0.78}{0.175} & \scalebox{0.78}{0.117} & \scalebox{0.78}{0.161} & \scalebox{0.78}{0.153} & \scalebox{0.78}{0.255} & \scalebox{0.78}{0.434} & \scalebox{0.78}{0.463} & \scalebox{0.78}{0.084} & \scalebox{0.78}{0.183} & \scalebox{0.78}{0.088} & \scalebox{0.78}{0.186} & \scalebox{0.78}{0.059} & \scalebox{0.78}{0.158} & \scalebox{0.78}{0.094} & \scalebox{0.78}{0.215}\\
& \scalebox{0.78}{6} & \textcolor{red}{\textbf{\scalebox{0.78}{0.060}}} & \textcolor{red}{\textbf{\scalebox{0.78}{0.156}}} & \textcolor{blue}{\underline{\scalebox{0.78}{0.064}}} & \textcolor{blue}{\underline{\scalebox{0.78}{0.165}}} & \scalebox{0.78}{0.076} & \scalebox{0.78}{0.190} & \scalebox{0.78}{0.129} & \scalebox{0.78}{0.173} & \scalebox{0.78}{0.157} & \scalebox{0.78}{0.257} & \scalebox{0.78}{0.526} & \scalebox{0.78}{0.541} & \scalebox{0.78}{0.092} & \scalebox{0.78}{0.191} & \scalebox{0.78}{0.096} & \scalebox{0.78}{0.195} & \scalebox{0.78}{0.065} & \scalebox{0.78}{0.167} & \scalebox{0.78}{0.112} & \scalebox{0.78}{0.234}\\
& \scalebox{0.78}{12} & \textcolor{red}{\textbf{\scalebox{0.78}{0.075}}} & \textcolor{red}{\textbf{\scalebox{0.78}{0.172}}} & \textcolor{blue}{\underline{\scalebox{0.78}{0.079}}} & \textcolor{blue}{\underline{\scalebox{0.78}{0.182}}} & \scalebox{0.78}{0.168} & \scalebox{0.78}{0.232} & \scalebox{0.78}{0.165} & \scalebox{0.78}{0.214} & \scalebox{0.78}{0.173} & \scalebox{0.78}{0.273} & \scalebox{0.78}{0.436} & \scalebox{0.78}{0.485} & \scalebox{0.78}{0.109} & \scalebox{0.78}{0.207} & \scalebox{0.78}{0.112} & \scalebox{0.78}{0.212} & \scalebox{0.78}{0.087} & \scalebox{0.78}{0.184} & \scalebox{0.78}{0.154} & \scalebox{0.78}{0.276}\\
& \scalebox{0.78}{24} & \textcolor{red}{\textbf{\scalebox{0.78}{0.108}}} & \textcolor{red}{\textbf{\scalebox{0.78}{0.204}}} & \textcolor{blue}{\underline{\scalebox{0.78}{0.115}}} & \textcolor{blue}{\underline{\scalebox{0.78}{0.219}}} & \scalebox{0.78}{0.224} & \scalebox{0.78}{0.281} & \scalebox{0.78}{0.215} & \scalebox{0.78}{0.260} & \scalebox{0.78}{0.210} & \scalebox{0.78}{0.301} & \scalebox{0.78}{0.467} & \scalebox{0.78}{0.502} & \scalebox{0.78}{0.140} & \scalebox{0.78}{0.236} & \scalebox{0.78}{0.141} & \scalebox{0.78}{0.238} & \scalebox{0.78}{0.122} & \scalebox{0.78}{0.221} & \scalebox{0.78}{0.248} & \scalebox{0.78}{0.353}\\
\cmidrule(lr){2-22}
& \scalebox{0.78}{Avg} & \textcolor{red}{\textbf{\scalebox{0.78}{0.074}}} & \textcolor{red}{\textbf{\scalebox{0.78}{0.170}}} & \textcolor{blue}{\underline{\scalebox{0.78}{0.078}}} & \textcolor{blue}{\underline{\scalebox{0.78}{0.180}}} & \scalebox{0.78}{0.133} & \scalebox{0.78}{0.220} & \scalebox{0.78}{0.157} & \scalebox{0.78}{0.202} & \scalebox{0.78}{0.173} & \scalebox{0.78}{0.272} & \scalebox{0.78}{0.466} & \scalebox{0.78}{0.498} & \scalebox{0.78}{0.106} & \scalebox{0.78}{0.204} & \scalebox{0.78}{0.109} & \scalebox{0.78}{0.208} & \scalebox{0.78}{0.083} & \scalebox{0.78}{0.183} & \scalebox{0.78}{0.152} & \scalebox{0.78}{0.270}\\ \midrule
& \scalebox{0.78}{$1^{st}$ Count} & \textcolor{red}{\textbf{\scalebox{0.78}{14}}} & \textcolor{red}{\textbf{\scalebox{0.78}{16}}} & \scalebox{0.78}{0} & \scalebox{0.78}{0} & \scalebox{0.78}{0} & \scalebox{0.78}{0} & \scalebox{0.78}{0} & \scalebox{0.78}{0} & \scalebox{0.78}{0} & \scalebox{0.78}{0} & \scalebox{0.78}{0} & \scalebox{0.78}{0} & \scalebox{0.78}{0} & \scalebox{0.78}{0} & \scalebox{0.78}{0} & \scalebox{0.78}{0} & \textcolor{blue}{\underline{\scalebox{0.78}{8}}} & \textcolor{blue}{\underline{\scalebox{0.78}{5}}} & \scalebox{0.78}{0} & \scalebox{0.78}{0}\\
             \bottomrule
          \end{tabular}
       \end{small}
    \end{threeparttable}}
     \vspace{-3mm}
 \end{table*}

\begin{table*}[t]
   \vspace{-5mm}
    \caption{Full results of multivariate long-term forecasting with a fixed input length $L = 96$ for all datasets. Baseline results are from \citet{iTransformer}.
    The best results are highlighted in \textcolor{red}{\textbf{red}} and the second bests are marked in \textcolor{blue}{\underline{blue}}. }\label{tab:full_long_forecasting_results}
    \vskip 0.05in
    \centering
    \setlength{\tabcolsep}{0.75pt}
    \resizebox{0.90\textwidth}{!}{
    \begin{threeparttable}
    \begin{small}
    \renewcommand{\multirowsetup}{\centering}
    \setlength{\tabcolsep}{1pt}
    \begin{tabular}{c|c|cc|cc|cc|cc|cc|cc|cc|cc|cc|cc}
 \toprule
 \multicolumn{2}{c}{Model} & 
 \multicolumn{2}{c}{\rotatebox{0}{\scalebox{0.8}{\textbf{Gateformer}}}} &\multicolumn{2}{c}{\rotatebox{0}{\scalebox{0.8}{iTransformer}}} &\multicolumn{2}{c}{\rotatebox{0}{\scalebox{0.8}{PatchTST}}} &\multicolumn{2}{c}{\rotatebox{0}{\scalebox{0.8}{Crossformer}}} &\multicolumn{2}{c}{\rotatebox{0}{\scalebox{0.8}{FEDformer}}} &\multicolumn{2}{c}{\rotatebox{0}{\scalebox{0.8}{Autoformer}}} &\multicolumn{2}{c}{\rotatebox{0}{\scalebox{0.8}{Stationary}}} &\multicolumn{2}{c}{\rotatebox{0}{\scalebox{0.8}{TimesNet}}} &\multicolumn{2}{c}{\rotatebox{0}{\scalebox{0.8}{SCINet}}} &\multicolumn{2}{c}{\rotatebox{0}{\scalebox{0.8}{DLinear}}} \\
 
 \multicolumn{2}{c}{ } & \multicolumn{2}{c}{\scalebox{0.76}{(\textbf{Ours})}} & 
 \multicolumn{2}{c}{\scalebox{0.76}{\citeyearpar{iTransformer}}} & 
 \multicolumn{2}{c}{\scalebox{0.76}{\citeyearpar{PatchTST}}} & 
 \multicolumn{2}{c}{\scalebox{0.76}{\citeyearpar{Crossformer}}} & 
 \multicolumn{2}{c}{\scalebox{0.76}{\citeyearpar{fedformer}}} & 
 \multicolumn{2}{c}{\scalebox{0.76}{\citeyearpar{Autoformer}}} & 
 \multicolumn{2}{c}{\scalebox{0.76}{\citeyearpar{Stationary}}} & 
 \multicolumn{2}{c}{\scalebox{0.76}{\citeyearpar{Timesnet}}} & 
 \multicolumn{2}{c}{\scalebox{0.76}{\citeyearpar{SCINet}}} &
 \multicolumn{2}{c}{\scalebox{0.76}{\citeyearpar{DLinear}}} \\
 
 \cmidrule(lr){3-4}\cmidrule(lr){5-6}\cmidrule(lr){7-8}\cmidrule(lr){9-10}\cmidrule(lr){11-12}\cmidrule(lr){13-14}\cmidrule(lr){15-16}\cmidrule(lr){17-18}\cmidrule(lr){19-20}\cmidrule(lr){21-22}
 \multicolumn{2}{c}{Metric} & \scalebox{0.78}{MSE} & \scalebox{0.78}{MAE} & \scalebox{0.78}{MSE} & \scalebox{0.78}{MAE} & \scalebox{0.78}{MSE} & \scalebox{0.78}{MAE} & \scalebox{0.78}{MSE} & \scalebox{0.78}{MAE} & \scalebox{0.78}{MSE} & \scalebox{0.78}{MAE} & \scalebox{0.78}{MSE} & \scalebox{0.78}{MAE} & \scalebox{0.78}{MSE} & \scalebox{0.78}{MAE} & \scalebox{0.78}{MSE} & \scalebox{0.78}{MAE} & \scalebox{0.78}{MSE} & \scalebox{0.78}{MAE} & \scalebox{0.78}{MSE} & \scalebox{0.78}{MAE} \\
 \toprule
 
 \multirow{5}{*}{\rotatebox{90}{\scalebox{0.95}{Traffic}}} 
& \scalebox{0.78}{96} & \textcolor{red}{\textbf{\scalebox{0.78}{0.390}}} & \textcolor{red}{\textbf{\scalebox{0.78}{0.261}}} & \textcolor{blue}{\underline{\scalebox{0.78}{0.395}}} & \textcolor{blue}{\underline{\scalebox{0.78}{0.268}}} & \scalebox{0.78}{0.462} & \scalebox{0.78}{0.295} & \scalebox{0.78}{0.522} & \scalebox{0.78}{0.290} & \scalebox{0.78}{0.587} & \scalebox{0.78}{0.366} & \scalebox{0.78}{0.613} & \scalebox{0.78}{0.388} & \scalebox{0.78}{0.612} & \scalebox{0.78}{0.338} & \scalebox{0.78}{0.593} & \scalebox{0.78}{0.321} & \scalebox{0.78}{0.788} & \scalebox{0.78}{0.499} & \scalebox{0.78}{0.650} & \scalebox{0.78}{0.396} \\
& \scalebox{0.78}{192} & \textcolor{red}{\textbf{\scalebox{0.78}{0.409}}} & \textcolor{red}{\textbf{\scalebox{0.78}{0.270}}} & \textcolor{blue}{\underline{\scalebox{0.78}{0.417}}} & \textcolor{blue}{\underline{\scalebox{0.78}{0.276}}} & \scalebox{0.78}{0.466} & \scalebox{0.78}{0.296} & \scalebox{0.78}{0.530} & \scalebox{0.78}{0.293} & \scalebox{0.78}{0.604} & \scalebox{0.78}{0.373} & \scalebox{0.78}{0.616} & \scalebox{0.78}{0.382} & \scalebox{0.78}{0.613} & \scalebox{0.78}{0.340} & \scalebox{0.78}{0.617} & \scalebox{0.78}{0.336} & \scalebox{0.78}{0.789} & \scalebox{0.78}{0.505} & \scalebox{0.78}{0.598} & \scalebox{0.78}{0.370} \\
& \scalebox{0.78}{336} & \textcolor{red}{\textbf{\scalebox{0.78}{0.424}}} & \textcolor{red}{\textbf{\scalebox{0.78}{0.278}}} & \textcolor{blue}{\underline{\scalebox{0.78}{0.433}}} & \textcolor{blue}{\underline{\scalebox{0.78}{0.283}}} & \scalebox{0.78}{0.482} & \scalebox{0.78}{0.304} & \scalebox{0.78}{0.558} & \scalebox{0.78}{0.305} & \scalebox{0.78}{0.621} & \scalebox{0.78}{0.383} & \scalebox{0.78}{0.622} & \scalebox{0.78}{0.337} & \scalebox{0.78}{0.618} & \scalebox{0.78}{0.328} & \scalebox{0.78}{0.629} & \scalebox{0.78}{0.336} & \scalebox{0.78}{0.797} & \scalebox{0.78}{0.508} & \scalebox{0.78}{0.605} & \scalebox{0.78}{0.373} \\
& \scalebox{0.78}{720} & \textcolor{red}{\textbf{\scalebox{0.78}{0.459}}} & \textcolor{red}{\textbf{\scalebox{0.78}{0.296}}} & \textcolor{blue}{\underline{\scalebox{0.78}{0.467}}} & \textcolor{blue}{\underline{\scalebox{0.78}{0.302}}} & \scalebox{0.78}{0.514} & \scalebox{0.78}{0.322} & \scalebox{0.78}{0.589} & \scalebox{0.78}{0.328} & \scalebox{0.78}{0.626} & \scalebox{0.78}{0.382} & \scalebox{0.78}{0.660} & \scalebox{0.78}{0.408} & \scalebox{0.78}{0.653} & \scalebox{0.78}{0.355} & \scalebox{0.78}{0.640} & \scalebox{0.78}{0.350} & \scalebox{0.78}{0.841} & \scalebox{0.78}{0.523} & \scalebox{0.78}{0.645} & \scalebox{0.78}{0.394} \\ 
\cmidrule(lr){2-22}
& \scalebox{0.78}{Avg} & \textcolor{red}{\textbf{\scalebox{0.78}{0.421}}} & \textcolor{red}{\textbf{\scalebox{0.78}{0.276}}} & \textcolor{blue}{\underline{\scalebox{0.78}{0.428}}} & \textcolor{blue}{\underline{\scalebox{0.78}{0.282}}} & \scalebox{0.78}{0.481} & \scalebox{0.78}{0.304} & \scalebox{0.78}{0.550} & \scalebox{0.78}{0.304} & \scalebox{0.78}{0.610} & \scalebox{0.78}{0.376} & \scalebox{0.78}{0.628} & \scalebox{0.78}{0.379} & \scalebox{0.78}{0.624} & \scalebox{0.78}{0.340} & \scalebox{0.78}{0.620} & \scalebox{0.78}{0.336} & \scalebox{0.78}{0.804} & \scalebox{0.78}{0.509} & \scalebox{0.78}{0.625} & \scalebox{0.78}{0.383}\\ \midrule
\multirow{5}{*}{\rotatebox{90}{\scalebox{0.95}{Solar-Energy}}} 
& \scalebox{0.78}{96} & \textcolor{red}{\textbf{\scalebox{0.78}{0.182}}} & \textcolor{red}{\textbf{\scalebox{0.78}{0.222}}} & \textcolor{blue}{\underline{\scalebox{0.78}{0.203}}} & \textcolor{blue}{\underline{\scalebox{0.78}{0.237}}} & \scalebox{0.78}{0.234} & \scalebox{0.78}{0.286} & \scalebox{0.78}{0.310} & \scalebox{0.78}{0.331} & \scalebox{0.78}{0.242} & \scalebox{0.78}{0.342} & \scalebox{0.78}{0.884} & \scalebox{0.78}{0.711} & \scalebox{0.78}{0.215} & \scalebox{0.78}{0.249} & \scalebox{0.78}{0.250} & \scalebox{0.78}{0.292} & \scalebox{0.78}{0.237} & \scalebox{0.78}{0.344} & \scalebox{0.78}{0.290} & \scalebox{0.78}{0.378}\\
& \scalebox{0.78}{192} & \textcolor{red}{\textbf{\scalebox{0.78}{0.227}}} & \textcolor{red}{\textbf{\scalebox{0.78}{0.257}}} & \textcolor{blue}{\underline{\scalebox{0.78}{0.233}}} & \textcolor{blue}{\underline{\scalebox{0.78}{0.261}}} & \scalebox{0.78}{0.267} & \scalebox{0.78}{0.310} & \scalebox{0.78}{0.734} & \scalebox{0.78}{0.725} & \scalebox{0.78}{0.285} & \scalebox{0.78}{0.380} & \scalebox{0.78}{0.834} & \scalebox{0.78}{0.692} & \scalebox{0.78}{0.254} & \scalebox{0.78}{0.272} & \scalebox{0.78}{0.296} & \scalebox{0.78}{0.318} & \scalebox{0.78}{0.280} & \scalebox{0.78}{0.380} & \scalebox{0.78}{0.320} & \scalebox{0.78}{0.398} \\
& \scalebox{0.78}{336} & \textcolor{red}{\textbf{\scalebox{0.78}{0.242}}} & \textcolor{blue}{\underline{\scalebox{0.78}{0.274}}} & \textcolor{blue}{\underline{\scalebox{0.78}{0.248}}} & \textcolor{red}{\textbf{\scalebox{0.78}{0.273}}} & \scalebox{0.78}{0.290} & \scalebox{0.78}{0.315} & \scalebox{0.78}{0.750} & \scalebox{0.78}{0.735} & \scalebox{0.78}{0.282} & \scalebox{0.78}{0.376} & \scalebox{0.78}{0.941} & \scalebox{0.78}{0.723} & \scalebox{0.78}{0.290} & \scalebox{0.78}{0.296} & \scalebox{0.78}{0.319} & \scalebox{0.78}{0.330} & \scalebox{0.78}{0.304} & \scalebox{0.78}{0.389} & \scalebox{0.78}{0.353} & \scalebox{0.78}{0.415}\\
& \scalebox{0.78}{720} & \textcolor{red}{\textbf{\scalebox{0.78}{0.245}}} & \textcolor{blue}{\underline{\scalebox{0.78}{0.279}}} & \textcolor{blue}{\underline{\scalebox{0.78}{0.249}}} & \textcolor{red}{\textbf{\scalebox{0.78}{0.275}}} & \scalebox{0.78}{0.289} & \scalebox{0.78}{0.317} & \scalebox{0.78}{0.769} & \scalebox{0.78}{0.765} & \scalebox{0.78}{0.357} & \scalebox{0.78}{0.427} & \scalebox{0.78}{0.882} & \scalebox{0.78}{0.717} & \scalebox{0.78}{0.285} & \scalebox{0.78}{0.295} & \scalebox{0.78}{0.338} & \scalebox{0.78}{0.337} & \scalebox{0.78}{0.308} & \scalebox{0.78}{0.388} & \scalebox{0.78}{0.356} & \scalebox{0.78}{0.413}\\ 
\cmidrule(lr){2-22}
& \scalebox{0.78}{Avg} & \textcolor{red}{\textbf{\scalebox{0.78}{0.224}}} & \textcolor{red}{\textbf{\scalebox{0.78}{0.258}}} & \textcolor{blue}{\underline{\scalebox{0.78}{0.233}}} & \textcolor{blue}{\underline{\scalebox{0.78}{0.262}}} & \scalebox{0.78}{0.270} & \scalebox{0.78}{0.307} & \scalebox{0.78}{0.641} & \scalebox{0.78}{0.639} & \scalebox{0.78}{0.291} & \scalebox{0.78}{0.381} & \scalebox{0.78}{0.885} & \scalebox{0.78}{0.711} & \scalebox{0.78}{0.261} & \scalebox{0.78}{0.381} & \scalebox{0.78}{0.301} & \scalebox{0.78}{0.319} & \scalebox{0.78}{0.282} & \scalebox{0.78}{0.375} & \scalebox{0.78}{0.330} & \scalebox{0.78}{0.401}\\  \midrule
\multirow{5}{*}{\rotatebox{90}{\scalebox{0.95}{Electricity}}} 
& \scalebox{0.78}{96} & \textcolor{red}{\textbf{\scalebox{0.78}{0.146}}} & \textcolor{red}{\textbf{\scalebox{0.78}{0.238}}} & \textcolor{blue}{\underline{\scalebox{0.78}{0.148}}} & \textcolor{blue}{\underline{\scalebox{0.78}{0.240}}} & \scalebox{0.78}{0.181} & \scalebox{0.78}{0.270} & \scalebox{0.78}{0.219} & \scalebox{0.78}{0.314} & \scalebox{0.78}{0.193} & \scalebox{0.78}{0.308} & \scalebox{0.78}{0.201} & \scalebox{0.78}{0.317} & \scalebox{0.78}{0.169} & \scalebox{0.78}{0.273} & \scalebox{0.78}{0.168} & \scalebox{0.78}{0.272} & \scalebox{0.78}{0.247} & \scalebox{0.78}{0.345} & \scalebox{0.78}{0.197} & \scalebox{0.78}{0.282}\\
& \scalebox{0.78}{192} & \textcolor{red}{\textbf{\scalebox{0.78}{0.160}}} & \textcolor{red}{\textbf{\scalebox{0.78}{0.252}}} & \textcolor{blue}{\underline{\scalebox{0.78}{0.162}}} & \textcolor{blue}{\underline{\scalebox{0.78}{0.253}}} & \scalebox{0.78}{0.188} & \scalebox{0.78}{0.274} & \scalebox{0.78}{0.231} & \scalebox{0.78}{0.322} & \scalebox{0.78}{0.201} & \scalebox{0.78}{0.315} & \scalebox{0.78}{0.222} & \scalebox{0.78}{0.334} & \scalebox{0.78}{0.182} & \scalebox{0.78}{0.286} & \scalebox{0.78}{0.184} & \scalebox{0.78}{0.289} & \scalebox{0.78}{0.257} & \scalebox{0.78}{0.355} & \scalebox{0.78}{0.196} & \scalebox{0.78}{0.285}\\
& \scalebox{0.78}{336} & \textcolor{red}{\textbf{\scalebox{0.78}{0.176}}} & \textcolor{blue}{\underline{\scalebox{0.78}{0.270}}} & \textcolor{blue}{\underline{\scalebox{0.78}{0.178}}} & \textcolor{red}{\textbf{\scalebox{0.78}{0.269}}} & \scalebox{0.78}{0.204} & \scalebox{0.78}{0.293} & \scalebox{0.78}{0.246} & \scalebox{0.78}{0.337} & \scalebox{0.78}{0.214} & \scalebox{0.78}{0.329} & \scalebox{0.78}{0.231} & \scalebox{0.78}{0.338} & \scalebox{0.78}{0.200} & \scalebox{0.78}{0.304} & \scalebox{0.78}{0.198} & \scalebox{0.78}{0.300} & \scalebox{0.78}{0.269} & \scalebox{0.78}{0.369} & \scalebox{0.78}{0.209} & \scalebox{0.78}{0.301}\\
& \scalebox{0.78}{720} & \textcolor{blue}{\underline{\scalebox{0.78}{0.221}}} & \textcolor{red}{\textbf{\scalebox{0.78}{0.306}}} & \scalebox{0.78}{0.225} & \textcolor{blue}{\underline{\scalebox{0.78}{0.317}}} & \scalebox{0.78}{0.246} & \scalebox{0.78}{0.324} & \scalebox{0.78}{0.280} & \scalebox{0.78}{0.363} & \scalebox{0.78}{0.246} & \scalebox{0.78}{0.355} & \scalebox{0.78}{0.254} & \scalebox{0.78}{0.361} & \scalebox{0.78}{0.222} & \scalebox{0.78}{0.321} & \textcolor{red}{\textbf{\scalebox{0.78}{0.220}}} & \scalebox{0.78}{0.320} & \scalebox{0.78}{0.299} & \scalebox{0.78}{0.390} & \scalebox{0.78}{0.245} & \scalebox{0.78}{0.333}\\ 
\cmidrule(lr){2-22}
& \scalebox{0.78}{Avg} & \textcolor{red}{\textbf{\scalebox{0.78}{0.176}}} & \textcolor{red}{\textbf{\scalebox{0.78}{0.267}}} & \textcolor{blue}{\underline{\scalebox{0.78}{0.178}}} & \textcolor{blue}{\underline{\scalebox{0.78}{0.270}}} & \scalebox{0.78}{0.205} & \scalebox{0.78}{0.290} & \scalebox{0.78}{0.244} & \scalebox{0.78}{0.334} & \scalebox{0.78}{0.214} & \scalebox{0.78}{0.327} & \scalebox{0.78}{0.227} & \scalebox{0.78}{0.338} & \scalebox{0.78}{0.193} & \scalebox{0.78}{0.296} & \scalebox{0.78}{0.192} & \scalebox{0.78}{0.295} & \scalebox{0.78}{0.268} & \scalebox{0.78}{0.365} & \scalebox{0.78}{0.212} & \scalebox{0.78}{0.300}\\ \midrule
\multirow{5}{*}{\rotatebox{90}{\scalebox{0.95}{Weather}}} 
& \scalebox{0.78}{96} & \textcolor{blue}{\underline{\scalebox{0.78}{0.168}}} & \textcolor{red}{\textbf{\scalebox{0.78}{0.208}}} & \scalebox{0.78}{0.174} & \textcolor{blue}{\underline{\scalebox{0.78}{0.214}}} & \scalebox{0.78}{0.177} & \scalebox{0.78}{0.218} & \textcolor{red}{\textbf{\scalebox{0.78}{0.158}}} & \scalebox{0.78}{0.230} & \scalebox{0.78}{0.217} & \scalebox{0.78}{0.296} & \scalebox{0.78}{0.266} & \scalebox{0.78}{0.336} & \scalebox{0.78}{0.173} & \scalebox{0.78}{0.223} & \scalebox{0.78}{0.172} & \scalebox{0.78}{0.220} & \scalebox{0.78}{0.221} & \scalebox{0.78}{0.306} & \scalebox{0.78}{0.196} & \scalebox{0.78}{0.255}\\
& \scalebox{0.78}{192} & \textcolor{blue}{\underline{\scalebox{0.78}{0.216}}} & \textcolor{red}{\textbf{\scalebox{0.78}{0.253}}} & \scalebox{0.78}{0.221} & \textcolor{blue}{\underline{\scalebox{0.78}{0.254}}} & \scalebox{0.78}{0.225} & \scalebox{0.78}{0.259} & \textcolor{red}{\textbf{\scalebox{0.78}{0.206}}} & \scalebox{0.78}{0.277} & \scalebox{0.78}{0.276} & \scalebox{0.78}{0.336} & \scalebox{0.78}{0.307} & \scalebox{0.78}{0.367} & \scalebox{0.78}{0.245} & \scalebox{0.78}{0.285} & \scalebox{0.78}{0.219} & \scalebox{0.78}{0.261} & \scalebox{0.78}{0.261} & \scalebox{0.78}{0.340} & \scalebox{0.78}{0.237} & \scalebox{0.78}{0.296}\\
& \scalebox{0.78}{336} & \textcolor{blue}{\underline{\scalebox{0.78}{0.276}}} & \textcolor{red}{\textbf{\scalebox{0.78}{0.296}}} & \scalebox{0.78}{0.278} & \textcolor{red}{\textbf{\scalebox{0.78}{0.296}}} & \scalebox{0.78}{0.278} & \textcolor{blue}{\underline{\scalebox{0.78}{0.297}}} & \textcolor{red}{\textbf{\scalebox{0.78}{0.272}}} & \scalebox{0.78}{0.335} & \scalebox{0.78}{0.339} & \scalebox{0.78}{0.380} & \scalebox{0.78}{0.359} & \scalebox{0.78}{0.395} & \scalebox{0.78}{0.321} & \scalebox{0.78}{0.338} & \scalebox{0.78}{0.280} & \scalebox{0.78}{0.306} & \scalebox{0.78}{0.309} & \scalebox{0.78}{0.378} & \scalebox{0.78}{0.283} & \scalebox{0.78}{0.335}\\
& \scalebox{0.78}{720} & \textcolor{blue}{\underline{\scalebox{0.78}{0.352}}} & \textcolor{blue}{\underline{\scalebox{0.78}{0.348}}} & \scalebox{0.78}{0.358} & \textcolor{red}{\textbf{\scalebox{0.78}{0.347}}} & \scalebox{0.78}{0.354} & \textcolor{blue}{\underline{\scalebox{0.78}{0.348}}} & \scalebox{0.78}{0.398} & \scalebox{0.78}{0.418} & \scalebox{0.78}{0.403} & \scalebox{0.78}{0.428} & \scalebox{0.78}{0.419} & \scalebox{0.78}{0.428} & \scalebox{0.78}{0.414} & \scalebox{0.78}{0.410} & \scalebox{0.78}{0.365} & \scalebox{0.78}{0.359} & \scalebox{0.78}{0.377} & \scalebox{0.78}{0.427} & \textcolor{red}{\textbf{\scalebox{0.78}{0.345}}} & \scalebox{0.78}{0.381}\\ 
\cmidrule(lr){2-22}
& \scalebox{0.78}{Avg} & \textcolor{red}{\textbf{\scalebox{0.78}{0.253}}} & \textcolor{red}{\textbf{\scalebox{0.78}{0.276}}} & \textcolor{blue}{\underline{\scalebox{0.78}{0.258}}} & \textcolor{blue}{\underline{\scalebox{0.78}{0.278}}} & \scalebox{0.78}{0.259} & \scalebox{0.78}{0.281} & \scalebox{0.78}{0.259} & \scalebox{0.78}{0.315} & \scalebox{0.78}{0.309} & \scalebox{0.78}{0.360} & \scalebox{0.78}{0.338} & \scalebox{0.78}{0.382} & \scalebox{0.78}{0.288} & \scalebox{0.78}{0.314} & \scalebox{0.78}{0.259} & \scalebox{0.78}{0.287} & \scalebox{0.78}{0.292} & \scalebox{0.78}{0.363} & \scalebox{0.78}{0.265} & \scalebox{0.78}{0.317}\\ \midrule
\multirow{5}{*}{\rotatebox{90}{\scalebox{0.95}{Exchange}}} 
& \scalebox{0.78}{96} & \textcolor{red}{\textbf{\scalebox{0.78}{0.081}}} & \textcolor{red}{\textbf{\scalebox{0.78}{0.199}}} & \textcolor{blue}{\underline{\scalebox{0.78}{0.086}}} & \scalebox{0.78}{0.206} & \scalebox{0.78}{0.088} & \textcolor{blue}{\underline{\scalebox{0.78}{0.205}}} & \scalebox{0.78}{0.256} & \scalebox{0.78}{0.367} & \scalebox{0.78}{0.148} & \scalebox{0.78}{0.278} & \scalebox{0.78}{0.197} & \scalebox{0.78}{0.323} & \scalebox{0.78}{0.111} & \scalebox{0.78}{0.237} & \scalebox{0.78}{0.107} & \scalebox{0.78}{0.234} & \scalebox{0.78}{0.267} & \scalebox{0.78}{0.396} & \scalebox{0.78}{0.088} & \scalebox{0.78}{0.218}\\
& \scalebox{0.78}{192} & \textcolor{red}{\textbf{\scalebox{0.78}{0.167}}} & \textcolor{red}{\textbf{\scalebox{0.78}{0.293}}} & \scalebox{0.78}{0.177} & \textcolor{blue}{\underline{\scalebox{0.78}{0.299}}} & \textcolor{blue}{\underline{\scalebox{0.78}{0.176}}} & \textcolor{blue}{\underline{\scalebox{0.78}{0.299}}} & \scalebox{0.78}{0.470} & \scalebox{0.78}{0.509} & \scalebox{0.78}{0.271} & \scalebox{0.78}{0.315} & \scalebox{0.78}{0.300} & \scalebox{0.78}{0.369} & \scalebox{0.78}{0.219} & \scalebox{0.78}{0.335} & \scalebox{0.78}{0.226} & \scalebox{0.78}{0.344} & \scalebox{0.78}{0.351} & \scalebox{0.78}{0.459} & \textcolor{blue}{\underline{\scalebox{0.78}{0.176}}} & \scalebox{0.78}{0.315}\\
& \scalebox{0.78}{336} & \textcolor{blue}{\underline{\scalebox{0.78}{0.312}}} & \textcolor{blue}{\underline{\scalebox{0.78}{0.403}}} & \scalebox{0.78}{0.331} & \scalebox{0.78}{0.417} & \textcolor{red}{\textbf{\scalebox{0.78}{0.301}}} & \textcolor{red}{\textbf{\scalebox{0.78}{0.397}}} & \scalebox{0.78}{1.268} & \scalebox{0.78}{0.883} & \scalebox{0.78}{0.460} & \scalebox{0.78}{0.427} & \scalebox{0.78}{0.509} & \scalebox{0.78}{0.524} & \scalebox{0.78}{0.421} & \scalebox{0.78}{0.476} & \scalebox{0.78}{0.367} & \scalebox{0.78}{0.448} & \scalebox{0.78}{1.324} & \scalebox{0.78}{0.853} & \scalebox{0.78}{0.313} & \scalebox{0.78}{0.427}\\
& \scalebox{0.78}{720} & \textcolor{red}{\textbf{\scalebox{0.78}{0.744}}} & \textcolor{red}{\textbf{\scalebox{0.78}{0.650}}} & \scalebox{0.78}{0.847} & \textcolor{blue}{\underline{\scalebox{0.78}{0.691}}} & \scalebox{0.78}{0.901} & \scalebox{0.78}{0.714} & \scalebox{0.78}{1.767} & \scalebox{0.78}{1.068} & \scalebox{0.78}{1.195} & \scalebox{0.78}{0.695} & \scalebox{0.78}{1.447} & \scalebox{0.78}{0.941} & \scalebox{0.78}{1.092} & \scalebox{0.78}{0.769} & \scalebox{0.78}{0.964} & \scalebox{0.78}{0.746} & \scalebox{0.78}{1.058} & \scalebox{0.78}{0.797} & \textcolor{blue}{\underline{\scalebox{0.78}{0.839}}} & \scalebox{0.78}{0.695}\\ 
\cmidrule(lr){2-22}
& \scalebox{0.78}{Avg} & \textcolor{red}{\textbf{\scalebox{0.78}{0.326}}} & \textcolor{red}{\textbf{\scalebox{0.78}{0.386}}} & \scalebox{0.78}{0.360} & \textcolor{blue}{\underline{\scalebox{0.78}{0.403}}} & \scalebox{0.78}{0.367} & \scalebox{0.78}{0.404} & \scalebox{0.78}{0.940} & \scalebox{0.78}{0.707} & \scalebox{0.78}{0.519} & \scalebox{0.78}{0.429} & \scalebox{0.78}{0.613} & \scalebox{0.78}{0.539} & \scalebox{0.78}{0.461} & \scalebox{0.78}{0.454} & \scalebox{0.78}{0.416} & \scalebox{0.78}{0.443} & \scalebox{0.78}{0.750} & \scalebox{0.78}{0.626} & \textcolor{blue}{\underline{\scalebox{0.78}{0.354}}} & \scalebox{0.78}{0.414}\\ \midrule
\multirow{5}{*}{\rotatebox{90}{\scalebox{0.95}{ETTm1}}} 
& \scalebox{0.78}{96} & \textcolor{red}{\textbf{\scalebox{0.78}{0.320}}} & \textcolor{red}{\textbf{\scalebox{0.78}{0.360}}} & \scalebox{0.78}{0.334} & \scalebox{0.78}{0.368} & \textcolor{blue}{\underline{\scalebox{0.78}{0.329}}} & \textcolor{blue}{\underline{\scalebox{0.78}{0.367}}} & \scalebox{0.78}{0.404} & \scalebox{0.78}{0.426} & \scalebox{0.78}{0.379} & \scalebox{0.78}{0.419} & \scalebox{0.78}{0.505} & \scalebox{0.78}{0.475} & \scalebox{0.78}{0.386} & \scalebox{0.78}{0.398} & \scalebox{0.78}{0.338} & \scalebox{0.78}{0.375} & \scalebox{0.78}{0.418} & \scalebox{0.78}{0.438} & \scalebox{0.78}{0.345} & \scalebox{0.78}{0.372}\\
& \scalebox{0.78}{192} & \textcolor{red}{\textbf{\scalebox{0.78}{0.367}}} & \textcolor{red}{\textbf{\scalebox{0.78}{0.383}}} & \scalebox{0.78}{0.377} & \scalebox{0.78}{0.391} & \textcolor{red}{\textbf{\scalebox{0.78}{0.367}}} & \textcolor{blue}{\underline{\scalebox{0.78}{0.385}}} & \scalebox{0.78}{0.450} & \scalebox{0.78}{0.451} & \scalebox{0.78}{0.426} & \scalebox{0.78}{0.441} & \scalebox{0.78}{0.553} & \scalebox{0.78}{0.496} & \scalebox{0.78}{0.459} & \scalebox{0.78}{0.444} & \textcolor{blue}{\underline{\scalebox{0.78}{0.374}}} & \scalebox{0.78}{0.387} & \scalebox{0.78}{0.439} & \scalebox{0.78}{0.450} & \scalebox{0.78}{0.380} & \scalebox{0.78}{0.389}\\
& \scalebox{0.78}{336} & \textcolor{blue}{\underline{\scalebox{0.78}{0.406}}} & \textcolor{blue}{\underline{\scalebox{0.78}{0.411}}} & \scalebox{0.78}{0.426} & \scalebox{0.78}{0.420} & \textcolor{red}{\textbf{\scalebox{0.78}{0.399}}} & \textcolor{red}{\textbf{\scalebox{0.78}{0.410}}} & \scalebox{0.78}{0.532} & \scalebox{0.78}{0.515} & \scalebox{0.78}{0.445} & \scalebox{0.78}{0.459} & \scalebox{0.78}{0.621} & \scalebox{0.78}{0.537} & \scalebox{0.78}{0.495} & \scalebox{0.78}{0.464} & \scalebox{0.78}{0.410} & \textcolor{blue}{\underline{\scalebox{0.78}{0.411}}} & \scalebox{0.78}{0.490} & \scalebox{0.78}{0.485} & \scalebox{0.78}{0.413} & \scalebox{0.78}{0.413}\\
& \scalebox{0.78}{720} & \textcolor{blue}{\underline{\scalebox{0.78}{0459}}} & \textcolor{blue}{\underline{\scalebox{0.78}{0.447}}} & \scalebox{0.78}{0.491} & \scalebox{0.78}{0.459} & \textcolor{red}{\textbf{\scalebox{0.78}{0.454}}} & \textcolor{red}{\textbf{\scalebox{0.78}{0.439}}} & \scalebox{0.78}{0.666} & \scalebox{0.78}{0.589} & \scalebox{0.78}{0.543} & \scalebox{0.78}{0.490} & \scalebox{0.78}{0.671} & \scalebox{0.78}{0.561} & \scalebox{0.78}{0.585} & \scalebox{0.78}{0.516} & \scalebox{0.78}{0.478} & \scalebox{0.78}{0.450} & \scalebox{0.78}{0.595} & \scalebox{0.78}{0.550} & \scalebox{0.78}{0.474} & \scalebox{0.78}{0.453}\\ 
\cmidrule(lr){2-22}
& \scalebox{0.78}{Avg} & \textcolor{blue}{\underline{\scalebox{0.78}{0.388}}} & \textcolor{red}{\textbf{\scalebox{0.78}{0.400}}} & \scalebox{0.78}{0.407} & \scalebox{0.78}{0.410} & \textcolor{red}{\textbf{\scalebox{0.78}{0.387}}} & \textcolor{red}{\textbf{\scalebox{0.78}{0.400}}} & \scalebox{0.78}{0.513} & \scalebox{0.78}{0.496} & \scalebox{0.78}{0.448} & \scalebox{0.78}{0.452} & \scalebox{0.78}{0.588} & \scalebox{0.78}{0.517} & \scalebox{0.78}{0.481} & \scalebox{0.78}{0.456} & \scalebox{0.78}{0.400} & \textcolor{blue}{\underline{\scalebox{0.78}{0.406}}} & \scalebox{0.78}{0.485} & \scalebox{0.78}{0.481} & \scalebox{0.78}{0.403} & \scalebox{0.78}{0.407}\\ \midrule
\multirow{5}{*}{\rotatebox{90}{\scalebox{0.95}{ETTm2}}} 
& \scalebox{0.78}{96} & \textcolor{blue}{\underline{\scalebox{0.78}{0.176}}} & \textcolor{blue}{\underline{\scalebox{0.78}{0.260}}} & \scalebox{0.78}{0.180} & \scalebox{0.78}{0.264} & \textcolor{red}{\textbf{\scalebox{0.78}{0.175}}} & \textcolor{red}{\textbf{\scalebox{0.78}{0.259}}} & \scalebox{0.78}{0.287} & \scalebox{0.78}{0.366} & \scalebox{0.78}{0.203} & \scalebox{0.78}{0.287} & \scalebox{0.78}{0.255} & \scalebox{0.78}{0.339} & \scalebox{0.78}{0.192} & \scalebox{0.78}{0.274} & \scalebox{0.78}{0.187} & \scalebox{0.78}{0.267} & \scalebox{0.78}{0.286} & \scalebox{0.78}{0.377} & \scalebox{0.78}{0.193} & \scalebox{0.78}{0.292}\\
& \scalebox{0.78}{192} & \textcolor{blue}{\underline{\scalebox{0.78}{0.245}}} & \textcolor{blue}{\underline{\scalebox{0.78}{0.307}}} & \scalebox{0.78}{0.250} & \scalebox{0.78}{0.309} & \textcolor{red}{\textbf{\scalebox{0.78}{0.241}}} & \textcolor{red}{\textbf{\scalebox{0.78}{0.302}}} & \scalebox{0.78}{0.414} & \scalebox{0.78}{0.492} & \scalebox{0.78}{0.269} & \scalebox{0.78}{0.328} & \scalebox{0.78}{0.281} & \scalebox{0.78}{0.340} & \scalebox{0.78}{0.280} & \scalebox{0.78}{0.339} & \scalebox{0.78}{0.249} & \scalebox{0.78}{0.309} & \scalebox{0.78}{0.399} & \scalebox{0.78}{0.445} & \scalebox{0.78}{0.284} & \scalebox{0.78}{0.362}\\
& \scalebox{0.78}{336} & \textcolor{red}{\textbf{\scalebox{0.78}{0.305}}} & \textcolor{red}{\textbf{\scalebox{0.78}{0.343}}} & \textcolor{blue}{\underline{\scalebox{0.78}{0.311}}} & \textcolor{blue}{\underline{\scalebox{0.78}{0.348}}} & \textcolor{red}{\textbf{\scalebox{0.78}{0.305}}} & \textcolor{red}{\textbf{\scalebox{0.78}{0.343}}} & \scalebox{0.78}{0.597} & \scalebox{0.78}{0.542} & \scalebox{0.78}{0.325} & \scalebox{0.78}{0.366} & \scalebox{0.78}{0.339} & \scalebox{0.78}{0.372} & \scalebox{0.78}{0.334} & \scalebox{0.78}{0.361} & \scalebox{0.78}{0.321} & \scalebox{0.78}{0.351} & \scalebox{0.78}{0.637} & \scalebox{0.78}{0.591} & \scalebox{0.78}{0.369} & \scalebox{0.78}{0.427}\\
& \scalebox{0.78}{720} & \textcolor{blue}{\underline{\scalebox{0.78}{0.404}}} & \textcolor{red}{\textbf{\scalebox{0.78}{0.399}}} & \scalebox{0.78}{0.412} & \scalebox{0.78}{0.407} & \textcolor{red}{\textbf{\scalebox{0.78}{0.402}}} & \textcolor{blue}{\underline{\scalebox{0.78}{0.400}}} & \scalebox{0.78}{1.730} & \scalebox{0.78}{1.042} & \scalebox{0.78}{0.421} & \scalebox{0.78}{0.415} & \scalebox{0.78}{0.433} & \scalebox{0.78}{0.432} & \scalebox{0.78}{0.417} & \scalebox{0.78}{0.413} & \scalebox{0.78}{0.408} & \scalebox{0.78}{0.403} & \scalebox{0.78}{0.960} & \scalebox{0.78}{0.735} & \scalebox{0.78}{0.554} & \scalebox{0.78}{0.522}\\ 
\cmidrule(lr){2-22}
& \scalebox{0.78}{Avg} & \textcolor{blue}{\underline{\scalebox{0.78}{0.283}}} & \textcolor{blue}{\underline{\scalebox{0.78}{0.327}}} & \scalebox{0.78}{0.288} & \scalebox{0.78}{0.332} & \textcolor{red}{\textbf{\scalebox{0.78}{0.281}}} & \textcolor{red}{\textbf{\scalebox{0.78}{0.326}}} & \scalebox{0.78}{0.757} & \scalebox{0.78}{0.610} & \scalebox{0.78}{0.305} & \scalebox{0.78}{0.349} & \scalebox{0.78}{0.327} & \scalebox{0.78}{0.371} & \scalebox{0.78}{0.306} & \scalebox{0.78}{0.347} & \scalebox{0.78}{0.291} & \scalebox{0.78}{0.333} & \scalebox{0.78}{0.571} & \scalebox{0.78}{0.537} & \scalebox{0.78}{0.350} & \scalebox{0.78}{0.401}\\ \midrule
\multirow{5}{*}{\rotatebox{90}{\scalebox{0.95}{ETTh1}}} 
& \scalebox{0.78}{96} & \textcolor{blue}{\underline{\scalebox{0.78}{0.383}}} & \textcolor{red}{\textbf{\scalebox{0.78}{0.398}}} & \scalebox{0.78}{0.386} & \scalebox{0.78}{0.405} & \scalebox{0.78}{0.414} & \scalebox{0.78}{0.419} & \scalebox{0.78}{0.423} & \scalebox{0.78}{0.448} & \textcolor{red}{\textbf{\scalebox{0.78}{0.376}}} & \scalebox{0.78}{0.419} & \scalebox{0.78}{0.449} & \scalebox{0.78}{0.459} & \scalebox{0.78}{0.513} & \scalebox{0.78}{0.491} & \scalebox{0.78}{0.384} & \scalebox{0.78}{0.402} & \scalebox{0.78}{0.654} & \scalebox{0.78}{0.599} & \scalebox{0.78}{0.386} & \textcolor{blue}{\underline{\scalebox{0.78}{0.400}}}\\
& \scalebox{0.78}{192} & \textcolor{blue}{\underline{\scalebox{0.78}{0.433}}} & \textcolor{blue}{\underline{\scalebox{0.78}{0.435}}} & \scalebox{0.78}{0.441} & \scalebox{0.78}{0.436} & \scalebox{0.78}{0.460} & \scalebox{0.78}{0.445} & \scalebox{0.78}{0.471} & \scalebox{0.78}{0.474} & \textcolor{red}{\textbf{\scalebox{0.78}{0.420}}} & \scalebox{0.78}{0.448} & \scalebox{0.78}{0.500} & \scalebox{0.78}{0.482} & \scalebox{0.78}{0.534} & \scalebox{0.78}{0.504} & \scalebox{0.78}{0.436} & \textcolor{red}{\textbf{\scalebox{0.78}{0.429}}} & \scalebox{0.78}{0.719} & \scalebox{0.78}{0.631} & \scalebox{0.78}{0.437} & \scalebox{0.78}{0.432}\\
& \scalebox{0.78}{336} & \textcolor{blue}{\underline{\scalebox{0.78}{0.464}}} & \textcolor{red}{\textbf{\scalebox{0.78}{0.452}}} & \scalebox{0.78}{0.487} & \textcolor{blue}{\underline{\scalebox{0.78}{0.458}}} & \scalebox{0.78}{0.501} & \scalebox{0.78}{0.466} & \scalebox{0.78}{0.570} & \scalebox{0.78}{0.546} & \textcolor{red}{\textbf{\scalebox{0.78}{0.459}}} & \scalebox{0.78}{0.465} & \scalebox{0.78}{0.521} & \scalebox{0.78}{0.496} & \scalebox{0.78}{0.588} & \scalebox{0.78}{0.535} & \scalebox{0.78}{0.491} & \scalebox{0.78}{0.469} & \scalebox{0.78}{0.778} & \scalebox{0.78}{0.659} & \scalebox{0.78}{0.481} & \scalebox{0.78}{0.459}\\
& \scalebox{0.78}{720} & \textcolor{red}{\textbf{\scalebox{0.78}{0.485}}} & \textcolor{red}{\textbf{\scalebox{0.78}{0.484}}} & \scalebox{0.78}{0.503} & \scalebox{0.78}{0.491} & \textcolor{blue}{\underline{\scalebox{0.78}{0.500}}} & \textcolor{blue}{\underline{\scalebox{0.78}{0.488}}} & \scalebox{0.78}{0.653} & \scalebox{0.78}{0.621} & \scalebox{0.78}{0.506} & \scalebox{0.78}{0.507} & \scalebox{0.78}{0.514} & \scalebox{0.78}{0.512} & \scalebox{0.78}{0.643} & \scalebox{0.78}{0.616} & \scalebox{0.78}{0.521} & \scalebox{0.78}{0.500} & \scalebox{0.78}{0.836} & \scalebox{0.78}{0.699} & \scalebox{0.78}{0.519} & \scalebox{0.78}{0.516}\\
\cmidrule(lr){2-22}
& \scalebox{0.78}{Avg} & \textcolor{blue}{\underline{\scalebox{0.78}{0.441}}} & \textcolor{red}{\textbf{\scalebox{0.78}{0.442}}} & \scalebox{0.78}{0.454} & \textcolor{blue}{\underline{\scalebox{0.78}{0.447}}} & \scalebox{0.78}{0.469} & \scalebox{0.78}{0.454} & \scalebox{0.78}{0.529} & \scalebox{0.78}{0.522} & \textcolor{red}{\textbf{\scalebox{0.78}{0.440}}} & \scalebox{0.78}{0.460} & \scalebox{0.78}{0.496} & \scalebox{0.78}{0.487} & \scalebox{0.78}{0.570} & \scalebox{0.78}{0.537} & \scalebox{0.78}{0.458} & \scalebox{0.78}{0.450} & \scalebox{0.78}{0.747} & \scalebox{0.78}{0.647} & \scalebox{0.78}{0.456} & \scalebox{0.78}{0.452}\\ \midrule
\multirow{5}{*}{\rotatebox{90}{\scalebox{0.95}{ETTh2}}} 
& \scalebox{0.78}{96} & \scalebox{0.78}{0.306} & \scalebox{0.78}{0.351} & \textcolor{red}{\textbf{\scalebox{0.78}{0.297}}} & \textcolor{blue}{\underline{\scalebox{0.78}{0.349}}} & \textcolor{blue}{\underline{\scalebox{0.78}{0.302}}} & \textcolor{red}{\textbf{\scalebox{0.78}{0.348}}} & \scalebox{0.78}{0.745} & \scalebox{0.78}{0.584} & \scalebox{0.78}{0.358} & \scalebox{0.78}{0.397} & \scalebox{0.78}{0.346} & \scalebox{0.78}{0.388} & \scalebox{0.78}{0.476} & \scalebox{0.78}{0.458} & \scalebox{0.78}{0.340} & \scalebox{0.78}{0.374} & \scalebox{0.78}{0.707} & \scalebox{0.78}{0.621} & \scalebox{0.78}{0.333} & \scalebox{0.78}{0.387}\\
& \scalebox{0.78}{192} & \textcolor{red}{\textbf{\scalebox{0.78}{0.371}}} & \textcolor{red}{\textbf{\scalebox{0.78}{0.394}}} & \textcolor{blue}{\underline{\scalebox{0.78}{0.380}}} & \textcolor{blue}{\underline{\scalebox{0.78}{0.400}}} & \scalebox{0.78}{0.388} & \scalebox{0.78}{0.400} & \scalebox{0.78}{0.877} & \scalebox{0.78}{0.656} & \scalebox{0.78}{0.429} & \scalebox{0.78}{0.439} & \scalebox{0.78}{0.456} & \scalebox{0.78}{0.452} & \scalebox{0.78}{0.512} & \scalebox{0.78}{0.493} & \scalebox{0.78}{0.402} & \scalebox{0.78}{0.414} & \scalebox{0.78}{0.860} & \scalebox{0.78}{0.689} & \scalebox{0.78}{0.477} & \scalebox{0.78}{0.476}\\
& \scalebox{0.78}{336} & \textcolor{red}{\textbf{\scalebox{0.78}{0.415}}} & \textcolor{red}{\textbf{\scalebox{0.78}{0.430}}} & \scalebox{0.78}{0.428} & \textcolor{blue}{\underline{\scalebox{0.78}{0.432}}} & \textcolor{blue}{\underline{\scalebox{0.78}{0.426}}} & \scalebox{0.78}{0.433} & \scalebox{0.78}{1.043} & \scalebox{0.78}{0.731} & \scalebox{0.78}{0.496} & \scalebox{0.78}{0.487} & \scalebox{0.78}{0.482} & \scalebox{0.78}{0.486} & \scalebox{0.78}{0.552} & \scalebox{0.78}{0.551} & \scalebox{0.78}{0.452} & \scalebox{0.78}{0.452} & \scalebox{0.78}{1.000} & \scalebox{0.78}{0.744} & \scalebox{0.78}{0.594} & \scalebox{0.78}{0.541}\\
& \scalebox{0.78}{720} & \textcolor{red}{\textbf{\scalebox{0.78}{0.425}}} & \textcolor{red}{\textbf{\scalebox{0.78}{0.442}}} & \textcolor{blue}{\underline{\scalebox{0.78}{0.427}}} & \textcolor{blue}{\underline{\scalebox{0.78}{0.445}}} & \scalebox{0.78}{0.431} & \scalebox{0.78}{0.446} & \scalebox{0.78}{1.104} & \scalebox{0.78}{0.763} & \scalebox{0.78}{0.463} & \scalebox{0.78}{0.474} & \scalebox{0.78}{0.515} & \scalebox{0.78}{0.511} & \scalebox{0.78}{0.562} & \scalebox{0.78}{0.560} & \scalebox{0.78}{0.462} & \scalebox{0.78}{0.468} & \scalebox{0.78}{1.249} & \scalebox{0.78}{0.838} & \scalebox{0.78}{0.831} & \scalebox{0.78}{0.657}\\
\cmidrule(lr){2-22}
& \scalebox{0.78}{Avg} & \textcolor{red}{\textbf{\scalebox{0.78}{0.379}}} & \textcolor{red}{\textbf{\scalebox{0.78}{0.404}}} & \textcolor{blue}{\underline{\scalebox{0.78}{0.383}}} & \textcolor{blue}{\underline{\scalebox{0.78}{0.407}}} & \scalebox{0.78}{0.387} & \textcolor{blue}{\underline{\scalebox{0.78}{0.407}}} & \scalebox{0.78}{0.942} & \scalebox{0.78}{0.684} & \scalebox{0.78}{0.437} & \scalebox{0.78}{0.449} & \scalebox{0.78}{0.450} & \scalebox{0.78}{0.459} & \scalebox{0.78}{0.526} & \scalebox{0.78}{0.516} & \scalebox{0.78}{0.414} & \scalebox{0.78}{0.427} & \scalebox{0.78}{0.954} & \scalebox{0.78}{0.723} & \scalebox{0.78}{0.559} & \scalebox{0.78}{0.515}\\ \midrule

& \scalebox{0.78}{$1^{st}$ Count} & \textcolor{red}{\textbf{\scalebox{0.78}{28}}} & \textcolor{red}{\textbf{\scalebox{0.78}{33}}} & \scalebox{0.78}{1} & \scalebox{0.78}{5} & \textcolor{blue}{\underline{\scalebox{0.78}{10}}} & \textcolor{blue}{\underline{\scalebox{0.78}{9}}} & \scalebox{0.78}{3} & \scalebox{0.78}{0} & \scalebox{0.78}{4} & \scalebox{0.78}{0} & \scalebox{0.78}{0} & \scalebox{0.78}{0} & \scalebox{0.78}{0} & \scalebox{0.78}{0} & \scalebox{0.78}{1} & \scalebox{0.78}{1} & \scalebox{0.78}{0} & \scalebox{0.78}{0} & \scalebox{0.78}{1} & \scalebox{0.78}{0}\\
             \bottomrule
          \end{tabular}
       \end{small}
    \end{threeparttable}}
     \vspace{-3mm}
 \end{table*}
\end{document}